\PassOptionsToPackage{table}{xcolor}
\documentclass[11pt]{article}

\usepackage[final]{acl}

\usepackage{times}
\usepackage{latexsym}

\usepackage[T1]{fontenc}

\usepackage[utf8]{inputenc}
\usepackage{xcolor} 

\usepackage{microtype}
\usepackage{listings}
\usepackage{adjustbox}
\usepackage{inconsolata}

\usepackage{graphicx}
\usepackage{amsmath,amssymb}
\usepackage{algorithm,algorithmic}
\usepackage{booktabs}
\usepackage{tabularx}

\definecolor{purplecolor}{rgb}{0.85, 0.4, 1}

\usepackage{hyperref}
\usepackage{xcolor}
\usepackage{tcolorbox}
\usepackage{amsmath}
\usepackage{amsfonts}
\usepackage{nicefrac}
\usepackage{multirow}
\usepackage{svg}
\tcbuselibrary{breakable, skins}
\usepackage{booktabs}
\usepackage{url}
\usepackage{xspace}
\usepackage{float}
\usepackage{textcomp}
\usepackage{amssymb}
\usepackage{makecell}
\usepackage{threeparttable}
\usepackage{wasysym}
\usepackage{pifont}
\usepackage{subcaption}

\definecolor{darkgreen}{rgb}{0, 0.5, 0}

\title{ElementCheck: Complexity-Aware Long-Form Text Factuality Evaluation via Sentence Elements}

\author{
 \textbf{Xinming Wang\textsuperscript{1,2}},
 \textbf{Haoran Du\textsuperscript{3}},
 \textbf{Yi Chen\textsuperscript{1,2}},
 \textbf{Xueqing Chen\textsuperscript{2,4}},
\\
 \textbf{Jian Xu\textsuperscript{1}},
 \textbf{Hong-Ming Yang\textsuperscript{5}},
 \textbf{Han Hu\textsuperscript{5}},
\\
 \textbf{Yulong Chen\textsuperscript{6,7}},
 \textbf{Cheng-Lin Liu\textsuperscript{1,2}}
  \textbf{Xu-Yao Zhang\textsuperscript{1,\thanks{Corresponding authors.}}}
\\
\\
 \textsuperscript{1}Institute of Automation, Chinese Academy of Sciences,
 \textsuperscript{2}Zhongguancun Academy,
\\
 \textsuperscript{3}Fudan University,
 \textsuperscript{4} Computer Network Information Center, Chinese Academy of Sciences, 
\\
 \textsuperscript{5}Tencent,
  \textsuperscript{6}University of Aberdeen,
 \textsuperscript{7}University of Cambridge\\
{\texttt{wangxinming2024@ia.ac.cn}, ~\texttt{\{liucl, xyz\}@nlpr.ia.ac.cn}
}
}

\begin{document}
\maketitle

\begin{abstract}
Existing long-form factuality evaluation relies on the decompose-retrieve-verify pipeline. 
However, the pipeline suffers from noise from claim decomposition and fixed verification granularity, resulting in unreliable results.
We propose \textbf{ElementCheck}, a complexity-aware framework that verifies long-form outputs via sentence elements. 
Instead of uniformly decomposing sentences into atomic sub-claims, ElementCheck extracts entity pairs that are explicitly linked through verifiable connections in the original sentence as elements, and organizes these into an element graph.
The graph topology provides a structural signal for estimating sentence complexity, enabling direct verification for simple sentences and targeted element-level refinement and verification for complex ones. 
To support fine-grained evaluation, we construct a new benchmark \textbf{FastFact-Sent} by mapping isolated claims from FastFact-Bench back to their source sentences. Experiments on FastFact-Sent and two domain-specific benchmarks show ElementCheck consistently improves factuality verification across five backbone models while maintaining a favorable accuracy-cost trade-off. Further analyses demonstrate that complexity-aware verification reduces unnecessary re-verification and maintains stability across different backbones. The code is available at \href{https://github.com/gudehhh666/elementcheck.git}{Here}.

\end{abstract}

\section{Introduction}


\begin{figure}[ht]
  \begin{center}
\centerline{\includegraphics[width=\columnwidth]{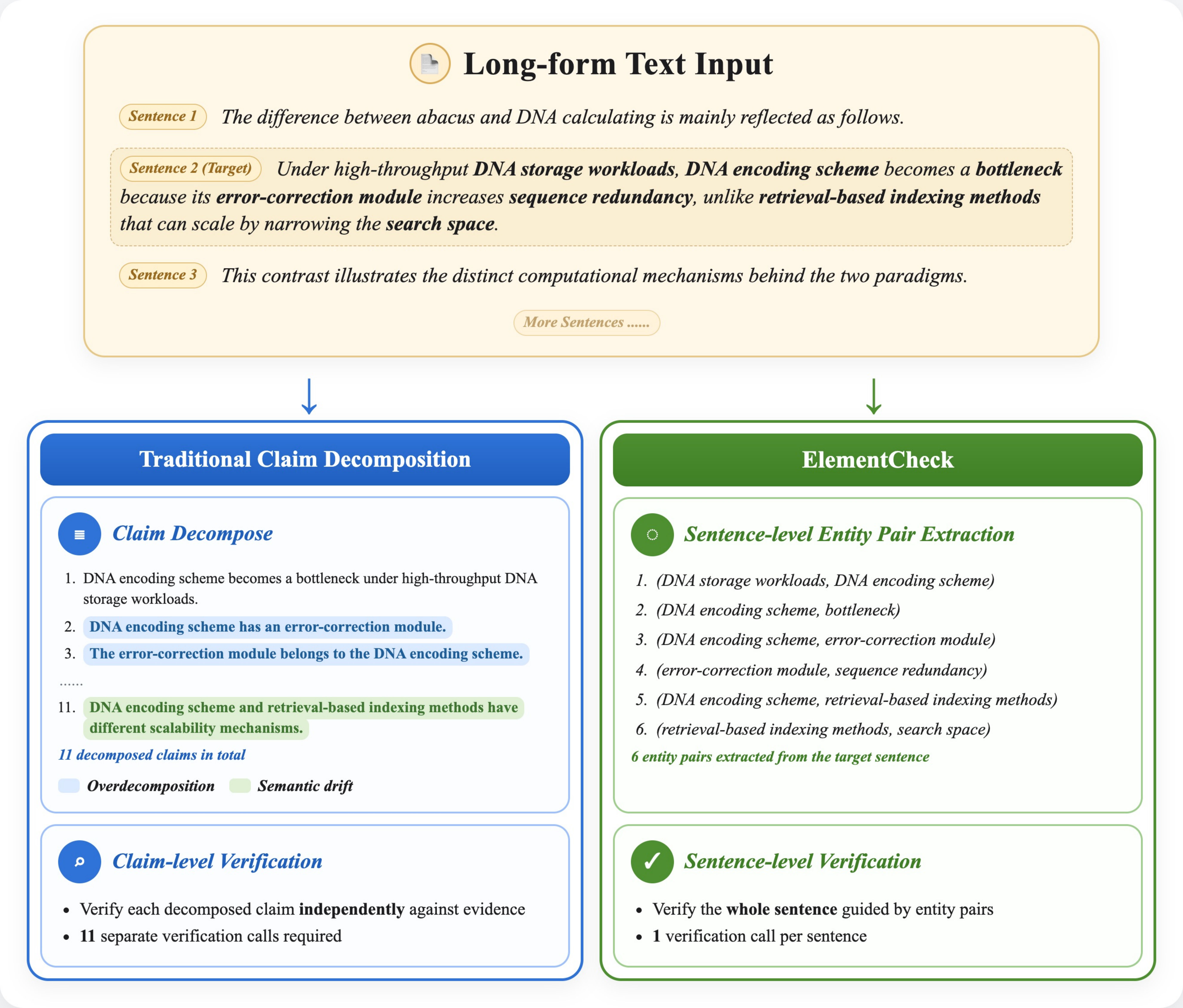}}
    \caption{Comparison between the ElementCheck pipeline and the Decompose-Retrieve-Verify pipeline.}
    \label{fig:main_compare}
  \end{center}
  \vspace{-24pt}
\end{figure}

With the advances of generation and reasoning abilities of large language models (LLMs)~\cite{chen2025recoverable,yu2024balancing}, models can handle complex text and generate long-form outputs consisting of multiple sentences~\citep{wang2025hitchhiker,bai2024longwriter}. 
Compared with short-form outputs, long-form text contains more factual claims with complex dependencies, where errors are difficult to localize and verify~\cite{wei_long-form_2024}.
Existing factuality evaluation typically relies on automatic fact-checking (AFC) systems based on a \textit{Decompose-Retrieve-Verify} pipeline~\cite{hu-etal-2025-decomposition}. 
These systems first decompose generated text into atomic claims, then retrieve supporting evidence, and finally verify the veracity of each claim~\citep{wei_long-form_2024,song-etal-2024-veriscore,min-etal-2023-factscore,wang2026mr}.

While being effective for short and well-bounded claims~\cite{sathe-etal-2020-automated,li-etal-2024-self}, existing systems suffer from two salient limitations for long-form text.
First, LLM-based claim decomposition often introduces noise. 
As shown in Figure~\ref{fig:main_compare}, the decomposition produces over-decomposed claims or claims affected by semantic drift. 
Such decomposition noise can compromise faithfulness or remove context needed for downstream verification~\citep{wanner-etal-2024-closer,mitra-etal-2025-factlens}.
Second, existing systems typically adopt a fixed verification granularity~\citep{min-etal-2023-factscore,song-etal-2024-veriscore}, ignoring differences in verification complexity.
Some claims (e.g., ``\textit{Retrieval-based indexing methods can scale}'') can be verified with minimal evidence or commonsense knowledge, while other claims require multi-hop or more fine-grained analysis over multiple pieces of evidence. 
Applying the same verification procedure can either lead to a computation waste on simple claims or fail to adequately verify complex ones~\citep{pan-etal-2023-fact, lu2025optimizing}.
These limitations call for a verification framework that preserves the original factual structure of long-form text while dynamically adapting verification granularity to claim complexity.

To address these limitations, we propose \textbf{ElementCheck}, a complexity-aware framework for adaptive long-form factuality verification via sentence elements. 
Instead of uniformly decomposing all sentences into atomic or sub-claims, ElementCheck first extracts entity pairs that are explicitly linked through verifiable connections as elements and constructs an element graph that serves as verification anchors.
The topology of this graph provides a non-learned, low-cost structural routing signal; after the graph has been extracted, the routing decision requires no additional model tokens. It does not replace sentence semantics, because the complete coreference-resolved sentence remains available to both verification paths.
Based on this signal, ElementCheck dynamically determines the verification granularity for different sentences. 
For structurally simple sentences, ElementCheck directly verifies them as complete units.
For structurally complex sentences, ElementCheck first verifies the individual factual elements within the graph to identify uncertain or unsupported regions.
When unverifiable elements are detected, it reconstructs refined claims from the corresponding local subgraphs and performs targeted re-verification.
In this way, ElementCheck transforms claim decomposition from a fixed step into a dynamic and adaptive refinement process, enabling finer-grained verification only when necessary while avoiding unnecessary costs for easy-verification cases.



A key obstacle in long-form factuality evaluation is the lack of fine-grained sentence-level resources for evaluating the reliability of long-form fact-checking pipelines. Existing studies, such as SAFE~\cite{wei_long-form_2024} and VeriScore~\cite{song-etal-2024-veriscore}, mainly evaluate the factuality of model responses, rather than the pipeline itself. Although FastFact-Bench~\cite{wan-etal-2025-fastfact} provides human-extracted claims at the response level, it does not explicitly retain their sentence-level provenance.
To this end, we construct FastFact-Sent from FastFact-Bench. We map each human-annotated claim back to its source sentence in the original response, thereby producing a fine-grained long-form factuality checking dataset with 5,020 factually verified sentences across 380 model responses.



We evaluate our ElementCheck against several factuality verification frameworks on FastFact-Sent, and two domain-specific benchmarks~\citep{chen2026catalystbench,chen2026an}.
Experimental results show that ElementCheck achieves the best average coverage-adjusted score across the five evaluated backbone groups and a favorable trade-off between verification performance and computational cost.
We further analyze the routing efficiency, refinement behavior, and cross-backbone stability of ElementCheck.
Analyses show that complexity-aware adaptive verification effectively allocates refinement effort based on structural complexity, substantially reducing unnecessary re-verification while maintaining strong verification performance across different models. 
We publicly release our \href{https://github.com/gudehhh666/elementcheck}{data and code}.

\section{Related Work}

\paragraph{Automated Fact-Checking}
Automated fact-checking (AFC) assesses the veracity of claims and is typically subdivided into claim detection \& extraction, evidence retrieval, and verdict prediction~\cite{akhtar2023multimodalautomatedfactcheckingsurvey}. As the spread of misinformation intensifies, AFC helps provide reliable, evidence-based verification~\cite{guo-etal-2022-survey}.
Evidence-based AFC has been extensively studied under controlled benchmarks, including FEVER~\cite{thorne-etal-2018-fever}, HoVer~\cite{jiang-etal-2020-hover}, WikiFactCheck~\cite{sathe-etal-2020-automated}, and ChartCheck~\cite{akhtar-etal-2024-chartcheck}.
To better reflect real-world verification, recent benchmarks retrieve evidence from the open web, such as MultiFC~\cite{augenstein-etal-2019-multifc} and AVeriTeC~\cite{schlichtkrull_averitec_2023}, with extensions to multimodal image--text claims in AVerImaTeC~\cite{cao2025averimatecdatasetautomaticverification}. SELFAR~\cite{vargas-etal-2024-improving} improves explainable fact-checking, and Ev2R~\cite{akhtar2025ev2revaluatingevidenceretrieval} further calls for explicit evaluation of evidence retrieval quality.
However, achieving robust end-to-end fact-checking remains a formidable challenge. Verification frameworks and models like AlignScore~\cite{zha-etal-2023-alignscore} and ANAH~\cite{ji-etal-2024-anah} suffer from significant performance degradation when encountering out-of-distribution verification tasks. Subsequent frameworks, such as MiniCheck~\cite{tang-etal-2024-minicheck}, SELF-CHECKER~\cite{li-etal-2024-self}, and DEFAME~\cite{braundefame}, explore multi-step pipelines to orchestrate the automated verification process. This fine-grained verification improves quality but comes with higher compute and latency~\cite{10.1145/3722212.3725098,nanekhan2025flashcheckexplorationefficientevidence}, preventing current fine-grained methods from scaling to long-form scenarios. However, these approaches primarily target single-claim verification; when applied to long-form generations, the number of elements grows rapidly, amplifying the trade-off between precision and efficiency, motivating dedicated long-form factuality evaluation.

\begin{figure*}[ht]
  \begin{center}
    \centerline{\includegraphics[width=\textwidth]{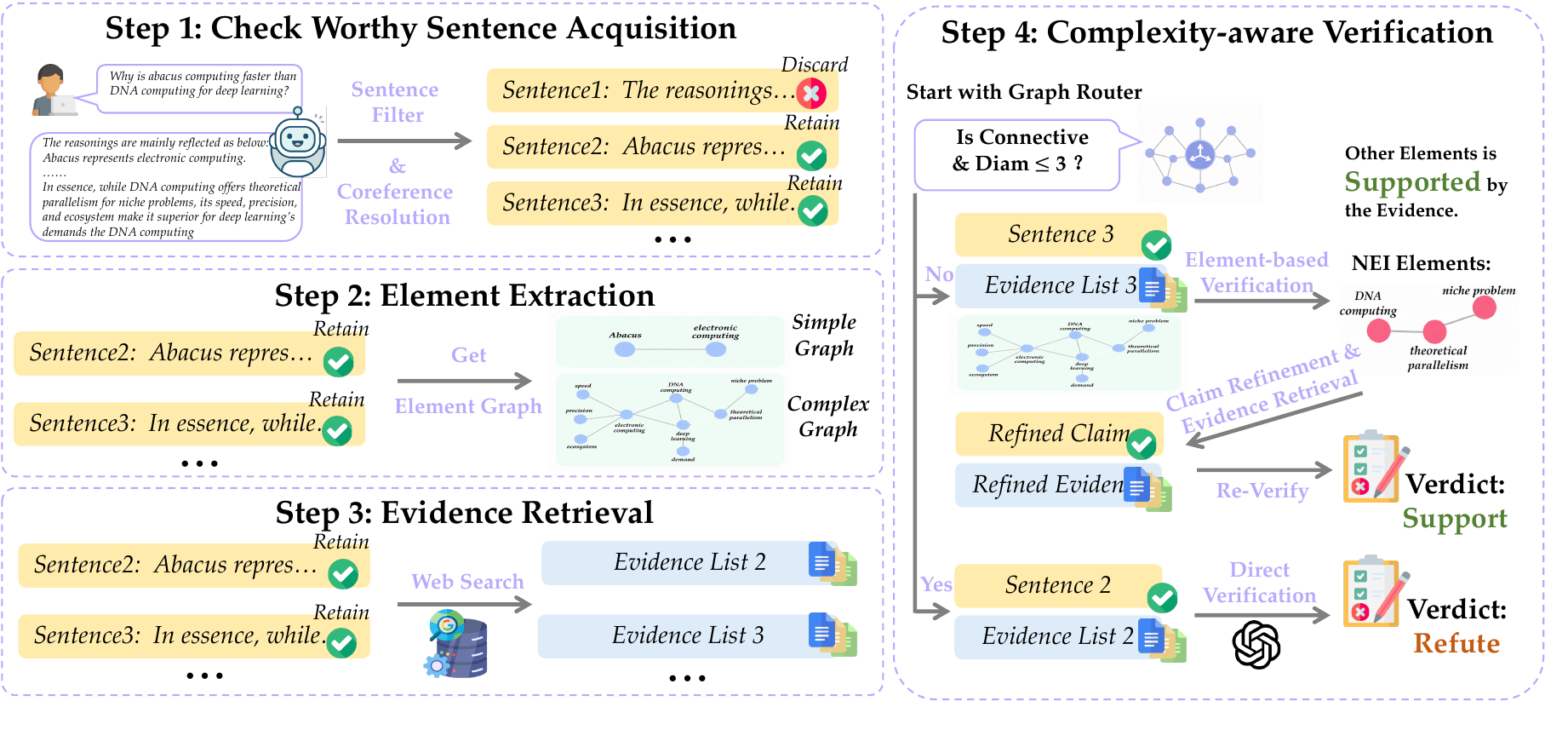}}
    \caption{
      The Overview of ElementCheck pipeline.
    }
    \label{fig:main_workflow}
  \end{center}
  \vspace{-24pt}
\end{figure*}

\paragraph{Long-Form Factuality Evaluation}
Given high claim density and mixed verifiability, FactScore~\cite{min-etal-2023-factscore} was the first to propose fact-checking long texts based on atomic claims. GraphCheck~\cite{chen2025graphcheckbreakinglongtermtext} also uses a graph, but under a different interface: it verifies predefined claims against a fixed corpus or grounding document and uses an extracted relation graph in the verification representation. ElementCheck instead acquires sentences from free-form long responses, retrieves open-web evidence, and uses an unlabeled entity-pair graph only as a routing signal while retaining the sentence as the semantic input. FactTool~\cite{chern2023factoolfactualitydetectiongenerative} enhances verification reliability by calling external retrieval and execution tools.
LongFact~\cite{wei_long-form_2024} exemplifies this paradigm via a search-augmented evaluator. Recent work improves efficiency by streamlining the pipeline, such as VeriFastScore~\cite{rajendhran2025verifastscorespeedinglongformfactuality} and FastFact~\cite{wan-etal-2025-fastfact}.
Meanwhile, a growing body of work recognizes the impact of claim selection, decomposition granularity, and context preservation on long-form fact-checking. Factcheck-Bench~\cite{wang-etal-2024-factcheck} evaluates check-worthiness, decomposition, decontextualization, retrieval, and verification as separate subtasks. A Closer Look at Claim Decomposition~\cite{wanner-etal-2024-closer} and FactLens~\cite{mitra-etal-2025-factlens} show that decomposition faithfulness and granularity affect downstream verification. Claimify~\cite{metropolitansky2025effectiveextractionevaluationfactual} resolves linguistic ambiguities during decomposition, and DNDSCORE~\cite{wanner-etal-2025-dndscore} examines the interaction between decomposition and decontextualization.
VERIFACT~\cite{liu-etal-2025-verifact} optimizes fact extraction by optimizing the context. However, these operations are still claim-centric, which largely introduces new overhead costs. In contrast, this paper focuses on sentence-level factual elements as verification anchors, reducing reliance on full claim decomposition and thereby mitigating decomposition-induced instability while controlling additional verification cost.

\section{Method}

The overview of ElementCheck is shown in Figure~\ref{fig:main_workflow}. In this section, we elaborate on the complete ElementCheck pipeline, including Check-worthy Sentence Acquisition, Element Extraction, Evidence Retrieval, and Complexity-aware Verification.

\subsection{Check-worthy Sentence Acquisition}

ElementCheck adopts a sentence-as-claim strategy rather than starting from atomic claim extraction. Check-worthiness detection is an established stage in automatic fact-checking~\cite{wang-etal-2024-factcheck}; here, we operationalize it for sentence-level verification rather than introduce a new task. The key observation is that a check-worthy sentence is often sufficiently self-contained to support a factuality verdict, while its decomposed sub-claims remain semantically subsumed by the sentence itself. This design mitigates decomposition-induced instability, such as over-decomposition and semantic drift, while keeping the original sentence context available for verification.

Formally, given an input text with preliminary sentence sequence $\mathcal{S}=\{S_1, S_2, \dots, S_m\}$, we then apply a check-worthiness acquisition operator by an elaborately designed prompt $\textsc{CW}(\cdot)$ to obtain a refined sequence of factuality-checkable sentences:
\begin{equation}
    \mathcal{S}^* = \{s^*_1, s^*_2, \dots, s^*_n\} = \textsc{CW}(\mathcal{S}),
\end{equation}
where $n \leq m$ indicates the filtration of intrinsically unverifiable content, such as subjective comments, imperative sentences and navigational discourse. The sentence sequence is obtained through \verb|sat-12l-sm|~\cite{frohmann-etal-2024-segment} segmentation.

\subsection{Element Extraction}

The essence of fact-checking lies in verifying discrete factual units~\cite{metropolitansky2025effectiveextractionevaluationfactual}. To rigorously quantify this process, we define elements according to whether entities are explicitly linked by linguistic structures that jointly express a verifiable fact, rather than by surface-level co-occurrence.
Given a check-worthy sentence $s^*$, we extract a set of fact elements $\mathcal{E}_{s^*}$:
\begin{equation}
    \mathcal{E}_{s^*}
    =
    \{ (e_i, e_j) \mid e_i, e_j \in \mathcal{V}_{s^*},
    \ \textsc{Conn}(e_i,e_j)\},
\end{equation}
where $\mathcal{V}_{s^*}$ denotes the set of entities in $s^*$. The function $\textsc{Conn}(e_i,e_j;s^*)$ indicates that $e_i$ and $e_j$ are explicitly connected by a linguistically licensed factual relation in the sentence. Such connections include predicative relations, attributive relations, modifying relations, appositive or exemplification structures, temporal or spatial anchoring, and causal dependencies as shown in Figure~\ref{fig:main_compare}.

Based on the extracted elements, we construct an element graph:
\begin{equation}
G_{s^*} = (\mathcal{V}_{s^*}, \mathcal{E}_{s^*}),
\end{equation}
where nodes set $\mathcal{V}_{s^*}$ correspond to factual entities and edges set $\mathcal{E}_{s^*}$ indicate explicitly expressed factual connections between entity pairs.
We do not require each edge to be labeled with a specific relation type. Instead, an edge only marks that the two entities jointly participate in a verifiable fact licensed by the sentence, while the precise predicate, scope, modality, and attribution are preserved in the original sentence context.

Since the element graph reveals sentence-internal structure, it is similar to dependency parsing~\cite{nivre-etal-2020-universal} and Semantic Role Labeling (SRL)~\cite{carreras-marquez-2005-introduction}. While dependency parsing is syntax-centric and SRL is predicate-centric, the element graph is entity-centric and verification-oriented. For example, in ``A discovered B in 1970,'' an element edge $(A,B)$ marks the checkable discovery relation, whereas a dependency parse also represents token-level grammatical links involving the verb, preposition, and date.
We do not use dependency parsing or SRL as the primary representation because grammatical dependencies and predicate-role labels do not necessarily correspond to verifiable factual dependencies, and may introduce noise from parsing errors or the compression of attribution, modality, nominal modification, and causal structures into fixed schemas. In contrast, element graphs provide lightweight verification anchors while preserving the full information in the original sentence context.

\subsection{Evidence Retrieval}

We maintain the efficiency of the previous single-round setting by employing the Serper API~\footnote{https://serper.dev/}. Let $\mathcal{Ret}(\cdot)$ denote the search engine interface. For a query sentence $x$, the snippet-level evidence $\mathcal{E}_{\text{snp}}$ is defined as the top-$n$ search snippets:
\begin{equation}
    {E}_{} = \text{Top}_n(\mathcal{Ret}(x)),
\end{equation}
where $n=10$ is the default configuration.

To investigate the impact of evidence granularity, we introduce a document-level variant inspired by FastFact~\cite{wan-etal-2025-fastfact}. In this setting, we extract full-text content from the top-$k$ URLs ($k=5$) via Jina~\footnote{https://jina.ai/} and apply BM25~\cite{robertson1994okapi} to reranking for the top-5 snippet chunks.

\subsection{Complexity-aware Verification}

As shown in Algorithm~\ref{alg:verification}, ElementCheck performs factuality verification with adaptive granularity based on sentence complexity rather than applying claim decomposition uniformly to all sentences. The topology of each check-worthy sentence's element graph $G_{s}$ serves as a coarse structural proxy for sentence complexity. It is used only for routing: the complete coreference-resolved sentence remains available to both verification paths, preserving negation, quantities, modality, scope, and attribution that are not explicitly encoded by the unlabeled graph. Specifically, we use graph diameter $\text{Diam}(G_s)$ and graph connectivity $\text{Con}(G_s)$ to estimate whether a sentence should be verified holistically or refined into element-level verification. A large graph diameter indicates long-range factual dependencies, while a disconnected graph suggests that the sentence contains multiple independent factual components that may be hard to reliably judge by a single holistic verdict. Based on this, the graph router filters each sentence to an appropriate verification protocol:
\begin{equation}
\text{Ver}(s) =
\begin{cases}
&\text{Ver}_{\text{ele}}(\mathcal{E}_s, E),\\ & ~~ \text{if } \text{Diam}(G_s) > \delta ~\text{or}~ \neg \text{Con}(G_s), \\
&\text{Ver}_{\text{direct}}(s, E),  \qquad\text{otherwise}.
\end{cases}
\end{equation}
Here, $\delta$ is the graph-diameter threshold. As shown in Table~\ref{tab:graph-dist} on FastFact-Sent, 88.6\% of connected element graphs have diameter no larger than 3. We therefore use $\delta=3$ as a practical operating point; Table~\ref{tab:delta} reports its empirical sensitivity.

\paragraph{Direct Verification}
For structurally simple sentences, ElementCheck treats the sentence as a compact semantic unit and directly verifies it against the retrieved evidence set $E$. The target output is a ternary factuality label
$y \in \{\textsc{Support}, \textsc{Refute}, \textsc{NEI}\}$, following the standard factual verification setting introduced by FEVER~\cite{thorne-etal-2018-fever}. Formally, direct verification is defined as:
\begin{equation}
y_{\text{pred}}=\text{Ver}_{\text{direct}}(s,E).
\end{equation}
In practice, we concatenate the verifiable sentence $s$ with the retrieved evidence $E$ and prompt the LLM verifier to assign the final sentence-level label. 

\paragraph{Element-based Verification.}
For structurally complex sentences, ElementCheck performs element-based verification over the extracted fact elements $\mathcal{E}_s$. The full sentence $s$ is also supplied to the verifier and is omitted from the notation below only for compactness.
Instead of assigning a single holistic verdict to the entire sentence, the verifier predicts a local factuality label for each element:
\begin{equation}
    \mathcal{Y}=\{y_e \mid e \in \mathcal{E}_s\},
    y_e \in \{\textsc{Sup}, \textsc{Ref}, \textsc{NEI}\}.
\end{equation}
This allows ElementCheck to identify which factual connections are supported, refuted, or remain uncertain. At the end of element-based verification, ElementCheck aggregates the local verdicts into a sentence-level prediction. 
A refuted element leads to a \textsc{Refute} label, while a sentence whose elements are all supported is labeled as \textsc{Support}. 
When unresolved \textsc{NEI} elements remain, ElementCheck further refines these uncertain regions rather than directly assigning a final verdict. The verifier retains conflicting evidence as an internal diagnostic subtype; following the three-way VeriScore protocol, unresolved conflicts are mapped to \textsc{NEI} for sentence-level evaluation.

\begin{algorithm}[tb]
\small
  \vskip 0.1in
  \caption{Complexity-aware Verification}
  \label{alg:verification}
  \begin{algorithmic}
    \STATE {\bfseries Input:} Check-worthy sentence $s^*$, element graph $G_{s^*}=(\mathcal{V}, \mathcal{E}_{s^*})$, initial evidence $E$, retriever $\mathcal{R}$, threshold $\delta$
    \STATE {\bfseries Output:} Verdict $y \in \{\textsc{Support}, \textsc{Refute}, \textsc{NEI}\}$
    
    \STATE $d \gets \text{Diam}(G_{s^*})$; $isCon \gets \text{Con}(G_{s^*})$
    
    \IF{$d \le \delta$ {\bfseries and} $isCon$}
        \STATE {\bfseries return} $\text{Ver}_{\text{direct}}(s^*, E)$
    \ENDIF
    
    \STATE $\mathcal{Y} \gets \text{Ver}_{\text{ele}}(\mathcal{E}_{s^*}, E)$
    
    \IF{$\exists y_e \in \mathcal{Y}: y_e = \textsc{Refute}$}
        \STATE {\bfseries return} \textsc{Refute}
    \ENDIF
    
    \IF{$\forall y_e \in \mathcal{Y}: y_e = \textsc{Support}$}
        \STATE {\bfseries return} \textsc{Support}
    \ENDIF
    
    \STATE $\mathcal{E}_{\textsc{NEI}} \gets \{e \in \mathcal{E}_{s^*} \mid y_e = \textsc{NEI}\}$
    \STATE $\mathcal{G}_{\text{sub}} \gets \textsc{GroupSubgraphs}(\mathcal{E}_{\textsc{NEI}}, G_{s^*})$
    
    \FOR{\textbf{each} $G_j \in \mathcal{G}_{\text{sub}}$}
        \STATE $c'_j \gets \textsc{RefineClaim}(G_j, s^*)$
        \STATE $E'_j \gets \mathcal{R}(c'_j)$
        \STATE $y'_j \gets \text{Ver}_{\text{direct}}(c'_j, E'_j)$
    \ENDFOR
    
    \IF{$\exists y'_j: y'_j = \textsc{Refute}$}
        \STATE {\bfseries return} \textsc{Refute}
    \ELSIF{$\forall y'_j: y'_j = \textsc{Support}$}
        \STATE {\bfseries return} \textsc{Support}
    \ELSE
        \STATE {\bfseries return} \textsc{NEI}
    \ENDIF
  \end{algorithmic}
\end{algorithm}

\paragraph{Claim Refinement and Re-Verify.}
ElementCheck groups the \textsc{NEI} elements into local subgraphs and reconstructs them into contextualized refined claims.
Compared with isolated elements, the refined claims recover the necessary sentence context, including the relevant entities, modifiers, and dependencies around the uncertain factual connections.
For each refined claim, ElementCheck conducts a new evidence retrieval and then applies direct verification again with the newly retrieved evidence. 
This Re-Verify process allows the verifier to focus on the ambiguous parts of the sentence rather than repeating verification over the entire sentence.

\begin{table*}[htbp]
\vspace{-6pt}
  \centering
  \begin{small}
\resizebox{0.95\textwidth}{!}{%

    \begin{tabular}{l|cccccccccc|cc|cc|ccc} 
      \toprule
       \multirow{3}{*}{Method} & \multicolumn{10}{c|}{FastFact-Sent.} &
       \multicolumn{2}{c|}{\multirow{2}{*}{AdjuvantBench}}&
       \multicolumn{2}{c|}{\multirow{2}{*}{CatalystBench}}&
       \multicolumn{3}{c}{\multirow{2}{*}{Average}}\\
      & \multicolumn{2}{c}{ExpertQA}
      & \multicolumn{2}{c}{FCBench}
      & \multicolumn{2}{c}{Bio}
      & \multicolumn{2}{c}{HelloBench}
      & \multicolumn{2}{c|}{LongFact}
      & 
      & 
      &
      &
      \\
      & Cov & BAcc & Cov & BAcc & Cov & BAcc & Cov & BAcc & Cov & BAcc & Cov & BAcc & Cov & BAcc & Cov & BAcc& Overall\\
      \midrule
      \rowcolor{gray!15}
      \multicolumn{18}{c}{\textbf{\texttt{Other Methods}}} \\
      \midrule

VeriFastScore   & 86.3 & 63.0 & 90.3 & 66.0 & 79.5 & 71.0 & 85.3 & 67.6 & 82.8 & 66.7 & 92.3 & 53.8 & 86.6 & 54.9 
& 87.9 & 58.4&51.3\\
\midrule

\rowcolor{gray!15}
\multicolumn{18}{c}{\textbf{\texttt{ GPT-4o-mini}}} \\
\midrule

SAFE
& \textbf{98.8} & 54.9
& \textbf{98.3} & 55.7
& \textbf{98.5} & 58.7
& \textbf{98.8} & 54.1
& \textbf{98.3} & 53.0
& \textbf{90.6} & 52.6
& \textbf{94.5} & 51.6 
& \textbf{96.4} & 53.5
&51.6\\

VeriScore
& 76.8 & \textbf{70.1}
& 81.4 & \textbf{69.9}
& 91.5 & \underline{69.1}
& 83.2 & \textbf{70.3}
& 90.4 & \underline{68.0}
& 69.2 & 60.3
& 67.0 & \textbf{61.9} 
& 73.5 & \underline{64.5} 
&47.4\\

FastFact
& \underline{89.8} & \underline{68.7}
& 73.4 & 58.8
& 73.1 & 60.0
& 77.9 & 62.6
& 85.0 & 59.4
& 32.8 & \underline{61.6}
& 17.8 & 53.4 
 & 44.1 & 59.1
 &26.1\\

ElementCheck
& 72.7 & 67.2
& \underline{82.1} & \underline{64.0}
& \underline{91.7} & \textbf{74.3}
& \underline{84.2} & \underline{69.0}
& \underline{93.5} & \textbf{69.7}
& \underline{78.1} & \textbf{62.5}
& \underline{79.8} & \underline{60.4}
& \underline{80.9} & \textbf{64.8}
&\textbf{52.4}\\

\midrule
\rowcolor{gray!15}
\multicolumn{18}{c}{\textbf{\texttt{ GPT-5.1}}} \\
\midrule

VeriScore  
& \textbf{96.4} & 57.1 
& \textbf{91.5} & 59.8 
& \textbf{95.0} & 66.3 
& \underline{93.4} & 63.1 
& 83.3 & 61.0
& 51.2 & \underline{64.9}
& 25.3 & \textbf{65.1} 
& 55.9 & \underline{63.7}
&35.6\\

FastFact  
& 82.0 & \underline{64.0} 
& 76.5 & \underline{65.4} 
& 76.0 & \underline{73.8} 
& 47.5 & \textbf{74.3} 
& 33.5 & \textbf{72.2}
& \textbf{92.5} & \underline{64.6}
& \underline{71.2} & 54.0 
& \underline{73.8} & 62.6
&46.2\\

ElementCheck 
& \underline{83.3} & \textbf{69.8} 
& \underline{90.6} & \textbf{70.6} 
& \underline{94.1} & \textbf{76.3} 
& \textbf{95.4} & \underline{72.5} 
& \textbf{95.2} & \underline{69.3}
& \underline{89.0} & \textbf{66.4}
& \textbf{79.5} & \underline{56.7} 
& \textbf{86.8} & \textbf{65.2}
&\textbf{56.6}\\

\midrule
\rowcolor{gray!15}
\multicolumn{18}{c}{\textbf{\texttt{DeepSeek-V3.2}}} \\
\midrule
   VeriScore 
&\textbf{96.4}&\underline{64.0}
&\textbf{91.5}&\underline{63.0}
&\textbf{95.0}&60.6
&\textbf{93.4}&63.8
&\underline{83.3}&58.6
&\textbf{86.6}&59.4
&\underline{86.4}&\underline{52.0}
&\textbf{88.0}&\underline{58.3}
&51.3\\

FastFact 
&85.1&56.0
&90.6&54.7
&53.9&\underline{61.0}
&82.0&\underline{65.0}
&74.7&\underline{59.1}
&\textbf{86.6}&\underline{61.4}
&\textbf{87.9}&51.5
&84.6&56.6
&47.9\\

ElementCheck 
&\underline{85.3}&\textbf{69.8}
&\underline{91.3}&\textbf{70.8}
&\underline{87.6}&\textbf{74.7}
&\underline{90.8}&\textbf{72.1}
&\textbf{92.8}&\textbf{74.3}
&\underline{82.1}&\textbf{66.4}
&84.1&\textbf{62.1}
&\underline{85.3}&\textbf{67.7}
&\textbf{57.7}\\

\midrule
\rowcolor{gray!15}
\multicolumn{18}{c}{\textbf{\texttt{Gemini-2.5-Flash}}} \\
\midrule

VeriScore & 73.0 & 48.2 & 86.6 & 54.2 & \underline{95.4} & \underline{70.0} & \underline{84.1} & 60.1 & 85.1 & \underline{63.9} & 76.9 & \underline{72.1} & 57.8 & 53.1 & 73.2 & \underline{61.5} & 45.0 \\

FastFact & \underline{80.5} & \underline{61.2} & \textbf{94.7} & \underline{56.9} & 94.6 & 66.4 & 80.9 & \underline{66.5} & \textbf{90.7} & 59.8 & \textbf{91.8} & 58.8 & \textbf{78.8} & \textbf{58.8} & \textbf{86.3} & 59.9 & \underline{51.7} \\

ElementCheck & \textbf{88.7} & \textbf{65.6} & \underline{93.4} & \textbf{67.3} & \textbf{96.2} & \textbf{73.6} & \textbf{95.7} & \textbf{68.4} & \underline{89.4} & \textbf{70.2} & \underline{79.5} & \textbf{84.6} & \underline{62.7} & \underline{55.9} & \underline{78.3} & \textbf{69.8} & \textbf{54.7} \\

\midrule
\rowcolor{gray!15}
\multicolumn{18}{c}{\textbf{\texttt{Qwen3-235B-A22B-Instruct-2507}}} \\
\midrule
      
VeriScore 
& \textbf{96.4} & \underline{63.9} 
& \underline{90.5} & \underline{63.4} 
& \underline{93.0} & \underline{70.3} 
& \textbf{93.4} & \underline{67.5} 
& \underline{83.3} & \underline{68.5} 
& \underline{83.6} & \underline{61.1} 
& \underline{84.3} & 57.6 
& \underline{86.3} & \underline{62.7}
&54.1\\

FastFact 
& 39.5 & 57.0 
& 49.7 & 62.0 
& 31.9 & 60.2 
& 54.4 & 64.4 
& 58.6 & 63.5
& 45.8 & 56.5
& 44.2 & \underline{58.9} 
& 46.5 & 58.8
&27.3\\
ElementCheck 
& \underline{89.2} & \textbf{68.8} 
& \textbf{90.6} & \textbf{72.8} 
& \textbf{93.9} & \textbf{76.9} 
& \underline{89.4} & \textbf{72.5} 
& \textbf{93.8} & \textbf{71.2} 
& \textbf{95.0} & \textbf{74.0}
& \textbf{95.6} & \textbf{62.8} 
& \textbf{93.6} & \textbf{70.4}
&\textbf{65.9}\\

\bottomrule
    \end{tabular}
    }
  \end{small}
\caption{Comparison of ElementCheck with other methods across multiple datasets. Cov denotes verifiable-sentence coverage, BAcc denotes class-balanced accuracy, and Overall is computed as BAcc $\times$ Cov. For each backbone model, the best and second-best results are highlighted in \textbf{bold} and \underline{underlined}, respectively.}
    \label{tab: main-result}

    \vspace{-6pt}
\end{table*}

\section{Experiments Setup}

\paragraph{Baselines.}
To ensure a comprehensive evaluation, we benchmark our approach against four leading methods in long-form factuality evaluation: SAFE, VeriScore, FastFact, and VeriFastScore. Additionally, we compare with two claim-refinement baselines: DnDScore~\cite{wanner-etal-2025-dndscore}, which applies decontextualization and decomposition, and VeriFact~\cite{liu-etal-2025-verifact}, which enhances fact extraction with reference facts. For claim-refinement baselines, we evaluate on the same set of verifiable sentences. SAFE and VeriFastScore are fixed-configuration reference points; cross-backbone comparisons use the backbone-switchable VeriScore and FastFact pipelines. Detailed specifications are provided in Appendix~\ref{appendix:baselines}.

\paragraph{Datasets.}
To support sentence-level evaluation, we construct FastFact-Sent from FastFact-Bench. FastFact-Sent is obtained by remapping human-annotated verifiable claims in FastFact-Bench~\cite{wan-etal-2025-fastfact} back to their originating sentences in the original model responses. It contains 5,020 valid unique sentence alignments retained from 6,953 original claims.
To construct FastFact-Sent, we segment each response into sentences, retrieve candidate sentence-claim pairs using \verb|bge-m3| semantic similarity, and use GPT-5.1 to verify whether the candidate sentence explicitly entails the claim. Unreliable alignments are discarded after multiple attempts, yielding sentence-level factuality supervision while preserving the original response context. We validate the remapping quality through a human agreement study on 500 randomly sampled sentence-claim pairs. Two independent annotators achieve 97.4\% and 95.2\% agreement with GPT-5.1, respectively. An independent Qwen remapping check further tests mapping-stage reproducibility (Appendix~\ref{appendix:construction}).
We conduct experiments on 3 datasets: FastFact-Sent~\cite{wan-etal-2025-fastfact}, AdjuvantBench, and CatalystBench. FastFact-Sent aggregates model responses from diverse domains covering FactScore-Bio, Factcheck-Bench~\cite{wang-etal-2024-factcheck}, ExpertQA~\cite{malaviya-etal-2024-expertqa}, LongFact, and HelloBench~\cite{que2024hellobenchevaluatinglongtext}. Basic statistics of these datasets are reported in Table~\ref{tab:dataset}.

\begin{table}[htbp]
    \vspace{-4pt}
    \centering
    \small
\resizebox{\columnwidth}{!}{%
    \begin{tabular}{lcccc}
      \toprule
{Dataset}&Domain&Samples&$\text{Resp}_\text{word}$&$\text{Resp}_\text{sent}$\\
     \midrule
         FastFact-Sent&Aggregated &380&465.10& 30.8 \\
         AdjuvantBench &Science&69& 100.19& 5.65 \\
          CatalystBench&Science&160& 133.59& 7.86   \\
       \bottomrule
    \end{tabular}
    }
    \caption{Statistics of Experiment Datasets.}
    \label{tab:dataset}
    \vspace{-12pt}
\end{table}

\paragraph{Metrics}
Following the evaluation protocols of~\cite{hu-etal-2025-decomposition}, we cast the main experiment as a binary classification task by merging fine-grained labels into Supported ($\textsc{Sup}$) and Not Supported ($\textsc{NoSup}$), with the latter covering both Refute and NEI instances; we additionally report 3-class balanced accuracy (BAcc-3) by treating Support, Refute, and NEI separately. The scored unit is the ground-truth source sentence for every method. Sentence-level outputs are scored directly; claim/subclaim outputs are mapped to their highest-similarity source sentence with \verb|bge-m3| at 0.85, then aggregated with priority Refute, NEI, and Support. Conflicting evidence is treated as an internal subtype of NEI. \textbf{Cov} is the proportion of source sentences receiving a mapped verdict, and \textbf{BAcc} is computed on the covered subset to mitigate class imbalance.
We report \textbf{Overall} as $\text{Cov} \times \text{BAcc}$ to make uncovered sentences visible in the final comparison.
Details are provided in Appendix~\ref{appendix:details_metrics}.

\section{Experiments Results}

\subsection{Main Results}


\begin{table*}[t]
    \centering
    \small
    \begin{tabular}{lcccccc}
      \toprule
      Method & $\text{Avg}_\text{Claims} (r)\downarrow$ & $\text{Avg}_\text{Calls}(r)\downarrow$ & $\text{Avg}_\text{Search}(r)\downarrow$ & $\text{Prompt}_\text{Tokens}\downarrow$ & $\text{Response}_\text{Tokens}\downarrow$ & $\text{Overall}\uparrow$ \\
     \midrule
     SAFE& 94.03 &124.97&94.03&35907.07&23000.45 &51.6\\
     VeriScore& 33.61 &45.28&33.61&26845.05&\ \ 3915.75&47.6\\
     VeriFastScore& 28.25 &57.10&\underline{19.55}&-&-&\underline{51.3}\\
     FastFact& \textbf{17.30} &\textbf{22.69}&\textbf{\ \ 4.71}&\textbf{21336.56}&\textbf{\ \ 1488.68}&26.1\\
     ElementCheck& \underline{17.52} &\underline{26.03}&21.28&\underline{21626.75}& \underline{\ \ 3761.05}&\textbf{52.4}\\
       \bottomrule
    \end{tabular}
    \caption{Computational efficiency analysis of different methods conducted on GPT-4o-mini on the full FastFact-Sent. In ElementCheck, $\text{Avg}_\text{Claims}$ denotes the average number of sentences verified, whereas in FastFact, $\text{Avg}_\text{Search}$ specifically refers to the Jina API calls.}
    \label{tab:cost and efficiency}

    \vspace{-6pt}

\end{table*}

Table~\ref{tab: main-result} reports 2-class BAcc, Cov and the overall scores across five backbone models on 3 benchmarks. We set $\delta=3$ throughout. Due to the high computational cost of SAFE, we only report its results on GPT-4o-mini. Fine-grained 3-class results are provided in Appendix~\ref{app:3class-results}, where Support, Refute, and NEI are evaluated separately. We further compare with DnDScore and VeriFact in Appendix~\ref{app:refinement}. Main-table systems follow their released retrieval procedures and are therefore not fully retrieval-matched; Appendix~\ref{app:retrieval-control} reports configurations and a controlled FastFact backend substitution.

\paragraph{Accuracy at native and matched coverage.} ElementCheck achieves the best average Overall score in each backbone group while maintaining competitive sentence-level coverage and balanced accuracy, reaching 65.9\% under Qwen3-235B. To separate prediction quality from native coverage, Table~\ref{tab:matched-coverage} restricts ElementCheck, VeriScore, and FastFact to the same 2,468 source sentences. ElementCheck improves BAcc-3 by 0.0537 over VeriScore and 0.0917 over FastFact; paired bootstrap tests are significant after Holm correction. Its BAcc-2 is comparable to VeriScore ($-0.0059$, not significant) and exceeds FastFact by 0.0866. Full confidence intervals are in Appendix~\ref{app:matched-coverage}.

\begin{table}[t]
    \centering
    \small
    \begin{tabular}{lcc}
      \toprule
      Method & Matched BAcc-2 & Matched BAcc-3 \\
      \midrule
      VeriScore & \textbf{0.7439} & 0.5177 \\
      FastFact & 0.6514 & 0.4797 \\
      ElementCheck & 0.7380 & \textbf{0.5714} \\
      \bottomrule
    \end{tabular}
    \caption{Coverage-controlled results on 2,468 jointly covered FastFact-Sent sentences with GPT-4o-mini. Sentence outputs are scored directly; claim-level outputs use the common matcher, 0.85 threshold, and aggregation rules.}
    \label{tab:matched-coverage}
    \vspace{-8pt}
\end{table}

\paragraph{Consistent effectiveness across backbones.} Across 35 dataset--backbone cells, ElementCheck ranks first in 27 and within the top two in 34. This claim is supported by comparisons with backbone-switchable VeriScore and FastFact, while SAFE and VeriFastScore remain fixed references. Appendix~\ref{app:stability} further decouples extractor and verifier backbones; it indicates that verifier capacity contributes more variation than the choice of graph extractor, without establishing backbone invariance.

\paragraph{Results across evaluated domains.} ElementCheck maintains relatively low performance variance and strong average results across the evaluated general domains, such as HelloBench and Bio, and scientific domains, such as AdjuvantBench and CatalystBench. This is descriptive evidence across the tested datasets rather than a claim of out-of-distribution generalization.

\begin{table}[t]
    \centering
    \small
    \resizebox{\linewidth}{!}{
    \begin{tabular}{lccccc}
      \toprule
      \multirow{2}{*}{Retrieval}& \multirow{2}{*}{Re-Verify} & \multicolumn{4}{c}{FastFact-Sent.} \\
      &&Acc.&BAcc. & $\text{Rec}_\text{Sup}$ & $\text{Rec}_\text{NoSup}$\\
     \midrule
         Top-10 &  $\times$ & 79.25 & 55.31 & 94.52 & 16.09\\
         Top-20 &  $\times$ & 80.31 & 53.91 & \textbf{96.97} & 10.86\\
         Jina-5 &  $\times$ & 76.97 & 51.09 & 93.25 & 8.93\\
         Top-10 &  \checkmark & \textbf{81.74} & \textbf{70.45}& 88.93& \textbf{51.97}\\
         Top-20 &  \checkmark & {81.54} & 66.96 & 90.73 & 43.20 \\
         Jina-5 &  \checkmark & 80.41 & 65.59 & 89.74 & 41.43  \\
       \bottomrule
    \end{tabular}
    }
    \caption{Ablation study on evidence intensity and the re-verify modules using GPT-4o-mini in ElementCheck without the graph router. $\text{Rec}_\text{Sup}$ and $\text{Rec}_\text{NoSup}$ denote the recall respectively.}
    \label{tab: retrieval ablation}
    \vspace{-12pt}
\end{table}

\subsection{Cost and Efficiency}

As indicated in Table~\ref{tab:cost and efficiency}, ElementCheck achieves a competitive computational cost on GPT-4o-mini. Compared with search-heavy baselines, it reduces unnecessary calls while preserving sentence-level verification. FastFact remains cheaper because it aggressively prioritizes parametric knowledge, but it verifies fewer units and yields weaker verification quality. The relative cost ordering remains consistent across additional backbones in Appendix~\ref{app:overhead}.

\subsection{Ablations}
\label{sec:graph-guided}

The ablation experiments isolate the role of evidence retrieval, Re-Verify, and the Graph Router. 
We focus on whether additional verification for structurally complex sentences improves the performance of fact-checking and whether this benefit depends on element-graph complexity and when Re-Verify works.

\paragraph{Effect of Re-Verify and evidence retrieval.} Table~\ref{tab: retrieval ablation} compares retrieval settings under GPT-4o-mini while disabling the Graph Router. The results show that expanding the evidence pool alone does not reliably improve verification. In contrast, adding Re-Verify to the normal Top-10 evidence retrieval substantially improves balanced accuracy and especially strengthens recall for \textsc{NoSup} cases. We further observe that richer evidence retrieved from Jina does not necessarily improve verification accuracy. Instead, the expanded evidence context may contain more irrelevant or noisy information, which can interfere with the verifier's judgment.

\begin{table*}[t]
    \vspace{-6pt}
    \centering
    \small
    \begin{tabular}{cccccccccccc}
      \toprule
      Graph Router & \multicolumn{4}{c}{FastFact-Sent.} & \multicolumn{7}{c}{Utilities}\\
      &Acc.&BAcc. & $\text{Rec}_\text{Sup}$ & $\text{Rec}_\text{NoSup}$ &Number& Dir &Re& Acc$_\text{Dir}$ &BAcc$_\text{Dir}$ & Acc$_\text{Re}$ &BAcc$_\text{Re}$ \\
     \midrule
    \rowcolor{gray!15}
   \multicolumn{12}{c}{\textbf{\texttt{GPT-4o-mini}}} \\
$\times$ & 81.74  &70.45  & 88.93  & 51.97 &7060& 41.4& 58.6&
89.55 & 50.90 &75.50  & 72.67\\
\checkmark & 80.63  & 71.29  & 86.53  & 56.05 & 7010&78.6& 21.4& 
81.32  & 71.10&79.55  & 71.49\\
   \rowcolor{gray!15}
  \multicolumn{12}{c}{\textbf{\texttt{GPT-5.1}}} \\
$\times$ & 74.22  & 73.01  & 74.96  & 71.07 &8081&25.2   & 74.8 &
93.44  & 55.57 & 67.25  & 70.95\\
\checkmark & 69.48  & 73.39  & 66.99  & 79.79 &7987& 70.8   & 29.2& 
71.01  & 74.29 & 66.67  & 71.60\\
       \bottomrule
    \end{tabular}
        \caption{Ablation study of graph router on FastFact-Sent. \textbf{Sent.} denotes the number of extracted sentences, \textbf{Dir.} denotes the proportion of directly processed items, and \textbf{Re.} denotes the proportion of items undergoing re-verification. }
        \label{tab:graph ablation}
        \vspace{-10pt}

\end{table*}

\begin{figure}[ht]
  \begin{center}
  \centerline{\includegraphics[width=\columnwidth]{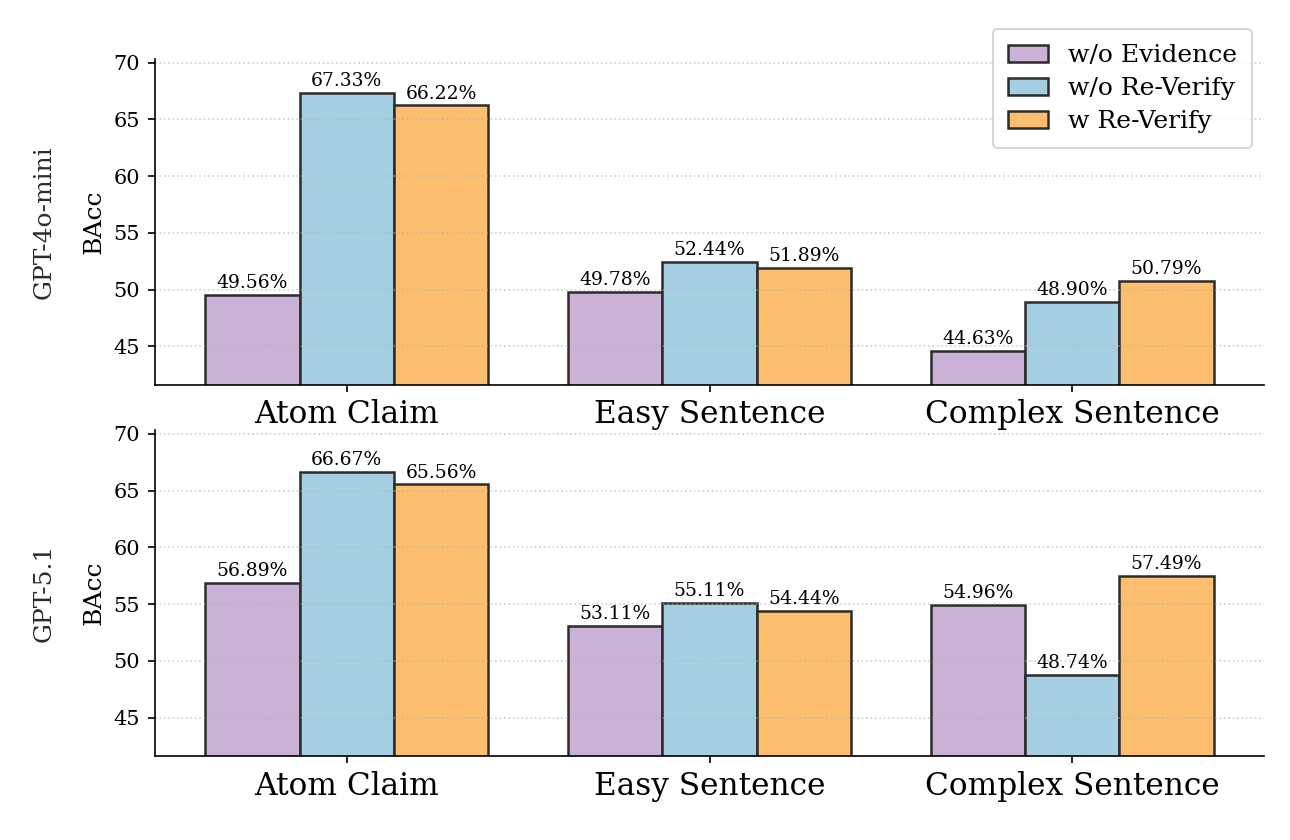}}
  \caption{Ablation analysis of verification strength across check-worthy sentences with different graph-structural complexity levels. The no-evidence setting denotes verification without evidence retrieval.}
  \label{fig:evidence ablations}
  \end{center}
    \vspace{-24pt}
\end{figure}

\paragraph{Effect depends on element-graph complexity.}
Figure~\ref{fig:evidence ablations} groups check-worthy sentences by graph-structural complexity. Re-Verify provides little benefit for structurally simple sentences, but it is clearly more useful for structurally complex sentences. This supports the central design choice of ElementCheck: stronger verification should be reserved for cases in which entity relations or qualifications make factuality difficult to determine.

\paragraph{Graph Router reduces unnecessary Re-Verify calls.} Table~\ref{tab:graph ablation} evaluates the routing module after fixing the retrieval setting. The Graph Router sends many more sentences to direct verification and correspondingly reduces the use of Re-Verify; under GPT-4o-mini, the re-verify's trigger rate drops from 58.6\% to 21.4\%. This reduction does not weaken balanced accuracy; instead, the performance remains stable and slightly improves across the tested backbones. 
\begin{table}[t]
    \centering
    \small
    \resizebox{\columnwidth}{!}{%
    \begin{tabular}{lccc}
      \toprule
      Router & BAcc-2 & BAcc-3 & Added tokens \\
      \midrule
      Random & 65.13 & 45.84 & 0 \\
      Sentence length & 65.48 & 45.37 & 0 \\
      Entity count & 65.67 & 46.29 & 0 \\
      GPT-4o-mini gate & 64.46 & 46.41 & 90M \\
      GPT-5 gate & 6573 & \textbf{46.86} & 2.50M \\
      Graph topology & \textbf{66.07} & 46.70 & 0 \\
      \bottomrule
    \end{tabular}
    }
    \caption{Matched-budget routing over precomputed GPT-4o-mini direct and flat-element verdicts on FastFact-Sent. Each router sends 37.13\% of sentences to the flat-element path; added tokens count only routing overhead.}
    \label{tab:matched-router}
    \vspace{-8pt}
\end{table}

\paragraph{Routing at a matched element-path budget.} Table~\ref{tab:matched-router} fixes the element-path fraction to ElementCheck's 37.13\% and reuses the same precomputed GPT-4o-mini direct and flat-element verdicts. Graph topology achieves the highest BAcc-2 and slightly exceeds entity count, sentence length, and random routing. GPT-5 gives the highest BAcc-3 by 0.0016 but consumes 2.50M additional router tokens; GPT-4o-mini gating is weaker and consumes 0.90M. Thus, the non-learned topology rule provides an effective accuracy--cost trade-off with zero additional router tokens after graph extraction, rather than uniformly dominating every learned gate. Details and the gate prompt are in Appendix~\ref{app:router-control}.
\begin{table}[ht]
    \centering
    \small
    \begin{tabular}{lcccc}
      \toprule
      $\delta$ & BAcc-3 & BAcc-2 & Cov & $\text{Rec}_\text{NoSup}$\\
     \midrule
         2 & 52.8 & 70.8 & 84.0 & 56.2 \\
         \textbf{3} & \textbf{53.8} & \textbf{71.3} & 86.5 & 56.1 \\
         4 & 53.6 & 71.1 & 86.3 & 53.4 \\
         5 & 53.0 & 70.1 & 86.6 & 51.6 \\
       \bottomrule
    \end{tabular}
    \caption{Threshold sensitivity of $\delta$ on FastFact-Sent with GPT-4o-mini.}
    \label{tab:delta}
    \vspace{-8pt}
\end{table}

\paragraph{Threshold sensitivity.} Table~\ref{tab:delta} reports the effect of varying $\delta$, which controls the boundary between structurally simple and complex element graphs. ElementCheck remains stable across the tested settings, with $\delta=3$ giving the best overall balance. This setting is consistent with the graph diameter distribution in Appendix~\ref{app:graph-stats}, but may require recalibration under substantial domain shift.

\begin{figure}[ht]
  \begin{center}
\centerline{\includegraphics[width=\columnwidth]{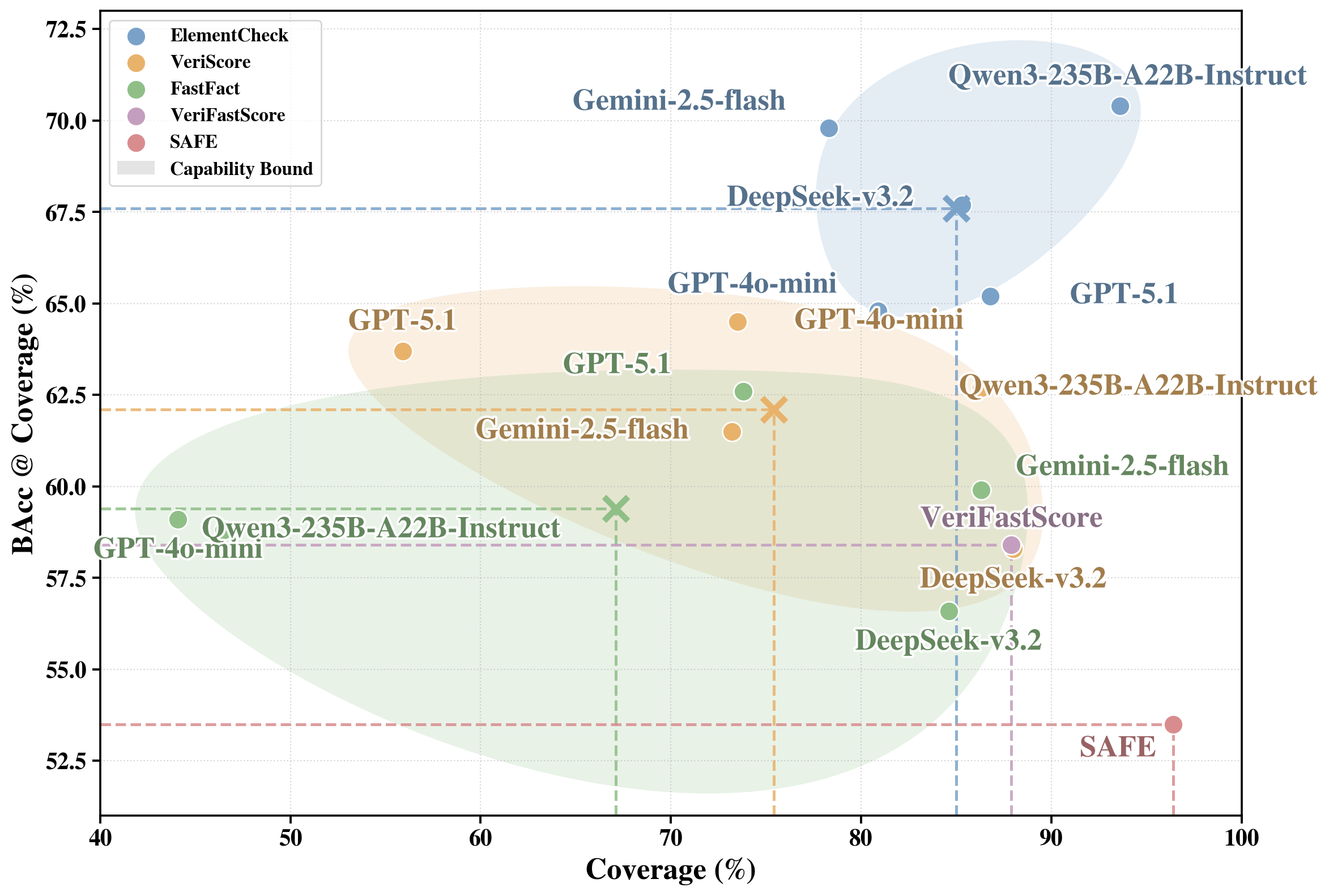}}
    \caption{Coverage and balanced-accuracy ranges across fact-checking methods and evaluated backbones.}
    \label{fig:main_coverage}
  \end{center}
  \vspace{-24pt}
\end{figure}



\section{Conclusion}


We proposed ElementCheck, a framework that grounds long-form factuality evaluation in sentence-level fact elements while preserving the complete sentence context. Across three benchmarks and five backbone groups, it achieves the best average coverage-adjusted score in each group. Matched-coverage and matched-routing-budget controls further show higher 3-class balanced accuracy than VeriScore and FastFact on the shared evaluation subset, and competitive routing accuracy without extra LLM-gate tokens. These results support sentence-preserving, complexity-aware routing as a practical alternative to applying claim decomposition uniformly.

\section*{Limitations} 

ElementCheck still has several limitations. First, its verifiable sentence acquisition stage depends on the capacity of the underlying backbone model, so weaker models may miss check-worthy content before verification begins. Second, the sentence-level formulation only partially addresses discourse-level factuality, including multi-paragraph evidence aggregation and broader causal inference. Finally, the current Re-Verify module reconstructs subgraph context with heuristic rules; future work could replace this step with more principled graph reasoning and evidence aggregation.

\section*{Ethical Considerations}

This work uses human validation only as a reliability check for our factuality verification process. The data used in this study are derived from public benchmarks and model-generated responses, and we do not intentionally collect private, personally identifiable, or sensitive user information.

\section*{Acknowledgments}

This work was supported in part by the CAS Project for Young Scientists in Basic Research under Grant YSBR-083, in part by Zhongguancun Academy under Project No.02012501, in part by CAAI-Tencent Rhino-Bird Open Research Fund (2026CAAI-Tencent-20).


\bibliography{example_paper}

\newpage
\appendix

\definecolor{lightgray}{gray}{0.95}
\definecolor{deepblue}{RGB}{70,130,180}
\definecolor{deepgray}{RGB}{119,136,153}
\lstdefinestyle{prompt}{
    basicstyle=\ttfamily\fontsize{7pt}{8pt}\selectfont,
    frame=none,
    breaklines=true,
    backgroundcolor=\color{lightgray},
    breakatwhitespace=true,
    breakindent=0pt,
    escapeinside={(*@}{@*)},
    numbers=none,
    numbersep=5pt,
    xleftmargin=5pt,
    aboveskip=2pt,
    belowskip=2pt,
}
\tcbset{
  aibox/.style={
    top=10pt,
    colback=white,
    center,
  }
}
\newtcolorbox{AIbox}[2][]{aibox, breakable, title=#2,#1}

\section{Experiments Details}

\subsection{Details of Baselines}
\label{appendix:baselines}
We summarize the key design choices of the long-form factuality baselines used in our study. All methods follow a broadly similar paradigm--decomposing a response into verifiable units, retrieving external evidence, and aggregating per-unit judgments.

\paragraph{SAFE} SAFE (Search-Augmented Factuality Evaluator) uses an LLM agent to decompose a long-form response into individual facts, issues web search queries for each fact, and then judges support or refutation based on retrieved results, producing an aggregate long-form factuality score.

\paragraph{VeriScore}
VeriScore is a reference-free factuality metric designed for long-form generations containing both verifiable and unverifiable content; it extracts verifiable claims and verifies them via retrieval, with empirical validation that its claim extraction is more sensible than competing approaches across diverse long-form tasks.

\paragraph{VeriFastScore}
VeriFastScore speeds up long-form factuality evaluation by training a verifier (Llama3.1-8B) on synthetic data to jointly extract and verify all verifiable claims in approximately a single model pass, given consolidated evidence, achieving strong correlation with the original VeriScore pipeline while substantially reducing runtime.

\paragraph{FastFact}
FaStFact is an efficiency-oriented long-form factuality evaluation pipeline that retains a decompose-and-verify structure while introducing chunk-level claim extraction and confidence-based pre-verification to reduce unnecessary search; it also emphasizes stronger evidence collection with document-level chunks to mitigate snippet insufficiency during verification.

\paragraph{DnDScore}
DnDScore studies claim refinement through decomposition and decontextualization. In our comparison, it is evaluated on the same verifiable sentences used by ElementCheck so that the comparison focuses on the effect of claim refinement rather than differences in upstream sentence selection.

\paragraph{VeriFact}
VeriFact improves fact extraction by refining claims with additional contextual or reference facts. We include it as a claim-refinement baseline under the same verifiable-sentence setting, and report the comparison in Appendix~\ref{app:refinement}.

\subsection{Details of Datasets}

\paragraph{FastFact-Bench.}
FastFact-Bench is an aggregated long-form factuality benchmark constructed for parallel evaluation of long-form factuality pipelines.
It comprises 400 long-form QA generations, covering five representative source benchmarks: FActScore-Bio, Factcheck-Bench, ExpertQA, LongFact, and HelloBench.
For each source benchmark, 80 prompts are sampled (total 400), spanning biography generation, general fact-seeking QA, domain-specific situational QA, concept/object explanation, and in-the-wild long-form writing/advice tasks.
The benchmark reports diverse prompt/response length statistics; averaged across four sampled LLMs, the aggregated responses are 465.1 words and 30.8 sentences on average.
Human annotations are provided to support both (i) claim extraction quality (e.g., revising ambiguous or missing claims) and (ii) claim verification outcomes, enabling evaluation reliability analyses beyond final scores.

\paragraph{CatalystBench.} CatalystBench is a multi-task benchmark designed for catalysis science that features a rigorous quality control process with high inter-annotator agreement (Cohen's Kappa of 0.75). We extract 160 instances from its expert-verified subset. Experts have identified the erroneous sentences in each answer, which we have marked as \textsc{Refute}; the others are marked as \textsc{Support} by default.

\paragraph{AdjuvantBench.}
AdjuvantBench evaluates capabilities in vaccine adjuvant research through open-ended QA and formal descriptions, all rigorously reviewed by domain experts. We utilize 69 examples from the benchmark's dedicated hallucination subset, directly adopting the provided expert adjudications as binary ground truth labels for sentence-level factuality, thus obviating the need for further manual annotation.

\subsection{Construction process of FastFact-Sent}
\label{appendix:construction}
To construct FastFact-Sent, a sentence-level dataset optimized for long-text fact-checking, we perform a rigorous remapping process that aligns atomic verifiable claims back to their originating sentences within the model responses. This process involves a two-stage pipeline: semantic retrieval ranking and model-based discrimination.

\paragraph{Semantic Retrieval Ranking} First, we segment the original long-text responses into individual sentences via \verb|sat-12l-sm|. To establish candidate links between claims and sentences, we utilize \verb|bge-m3| for semantic embedding.We encode both the extracted atomic claims and the segmented sentences into dense vectors. We calculate the cosine similarity between each claim-sentence pair. We initially select the sentence with the highest cosine similarity to the claim as a candidate.

\paragraph{Model-based Discrimination} Since semantic similarity alone may produce false positives, we use GPT-5.1 as a discriminator to refine the mapping. We input the claim and candidate sentence pairs identified in the previous step into the model. The model is asked to verify whether the candidate sentence explicitly contains or supports the atomic claim. Only sentence pairs that pass this discrimination check are retained. Otherwise, we select the sentence with the second-highest cosine similarity and perform the model discrimination again until a suitable sentence is found. Claims that fail the discrimination process after 5 attempts are discarded. 

The prompt for the discriminator is shown as Figure~\ref{fig:discriminator}.

\paragraph{Statistics of FastFact-Sent} 
Table~\ref{tab:fastfact_statistics} summarizes the detailed statistics of the constructed dataset. From the initial valid 380 response samples(20 responses have no claim annotated), we extract a total of 6,953 raw claims. After the semantic mapping and model-based filtering process, 5,908 claims and 5,020 unique verifiable sentences are retained as valid and mapped to their corresponding sentences.

\begin{table*}[htbp]
    \centering
    \small
    \begin{tabular}{ccccccc}
      \toprule
      Valid Samples & Valid Sentences & Raw Claims & Valid Claims & Support & Refute & NEI\\
     \midrule
     380 & 5020 & 6953 & 5908& 4098& 309& 613\\
       \bottomrule
    \end{tabular}
    \caption{Statistics of FastFact-Sent Dataset}
    \label{tab:fastfact_statistics}
    \vspace{-6pt}
\end{table*}

\paragraph{Inter-annotation agreement analysis}
To ensure the validity of the FastFact-Sent construction pipeline, we conducted a rigorous human evaluation to assess the consistency of GPT-5.1's judgments. We performed stratified random sampling to select 500 sentence-claim pairs. Two independent human annotators were tasked with the same objective as the model: determining whether the source sentence textually entails a given claim.

Both annotators are PhD students majoring in computer science with a strong educational background. The effective annotation duration per person is 10 hours, and both individuals have received corresponding remuneration for their work.

\begin{table*}[htbp]
    \centering
    \small
    \begin{tabular}{cccc}
      \toprule
      Annotator & Agreed & Disagreed & IAA \\
     \midrule
     Annotator 1 and GPT-5.1 & 487 & 13 & 97.4\% \\
     Annotator 2 and GPT-5.1 & 476 & 24 & 95.2\% \\
       \bottomrule
    \end{tabular}
    \caption{Human annotator agreement with GPT-5.1 on FastFact-Sent sentence-claim alignment.}
    \label{tab:human}
    \vspace{-6pt}
\end{table*}

As presented in Table \ref{tab:human}, the evaluation reveals a substantial alignment between the model and human judgments. Annotator 1 and Annotator 2 achieved agreement rates of 97.4\% and 95.2\% with GPT-5.1, respectively. We acknowledge that this high concordance is partially attributable to the explicit nature of the task, where the entailment relationship is largely syntactic and requires limited inference depth. This validation supports using GPT-5.1 for large-scale sentence-claim alignment, while the final factuality evaluation remains based on the dataset labels rather than the discriminator's own factuality judgments.

\paragraph{Independent mapping-stage reproducibility.}
We reran the mapping discriminator with Qwen3.5-397B-A17B while keeping sentence segmentation and candidate retrieval fixed. Qwen agreed with the retained GPT-5.1 claim decisions on 5,702 of 5,987 candidate decisions (95.24\%) and reproduced 4,996 of 5,110 sentence alignments (97.77\%). This check concerns only the claim-to-sentence mapping stage; it is not an end-to-end factuality result.

\paragraph{Case of FastFact-Sent}

Table~\ref{tab:sent_claim_label_agree} presents representative examples from FastFact-Sent. Each original sentence is aligned with one or more fact-checkable claims, and each claim is assigned a verification label according to the available evidence. The examples show that a single sentence may contain multiple factual units with different verification outcomes: some claims are supported by evidence, while others are labeled as NEI when the evidence does not explicitly establish the claim. The Agreement column indicates whether the sentence-claim alignment and the corresponding label are validated during annotation. These cases illustrate why sentence-level factuality supervision is necessary for long-form factuality evaluation, as response-level or sentence-level labels alone may obscure fine-grained factual distinctions within complex sentences.

\begin{table*}[htbp]
    \centering
    \small
    \setlength{\tabcolsep}{4pt}
    \renewcommand{\arraystretch}{1.15}
    \begin{tabularx}{\textwidth}{
        >{\raggedright\arraybackslash}p{0.4\textwidth}
        >{\raggedright\arraybackslash}p{0.33\textwidth}
        c c
    }
      \toprule
      Sentence & Claim & Label & Agreement \\
      \midrule

      \multirow{3}{0.4\textwidth}{\parbox[t]{0.4\textwidth}{\vspace{0pt}%
      \textbf{Carbohydrate Mouth Rinse}: Studies show swishing glucose solutions (without swallowing) tricks the brain into delaying fatigue (used by Chris Froome in the 2015 Tour de France). 
      \textbf{Caffeine}: Blocks adenosine receptors, reducing perceived effort (e.g., Paula Radcliffe's 2003 Marathon World Record, where she used caffeine gels).%
      }}
      & Paula Radcliffe used caffeine gels during her 2003 Marathon World Record performance. & NEI & True \\
      & Chris Froome used a carbohydrate mouth rinse in the 2015 Tour de France to delay fatigue. & NEI & True \\
      & Studies show swishing glucose solutions tricks the brain into delaying fatigue. & support & True \\
      \midrule

      Dr. Ross Tucker (sports scientist) argues that the CGM oversimplifies fatigue and doesn't fully account for peripheral muscle fatigue.
      & Dr. Ross Tucker argues that the Central Governor Model oversimplifies fatigue. & NEI & True \\
      \midrule

      Dr. Andrew Jones (University of Exeter) highlights that VO$_2$ max and lactate thresholds still play critical roles in performance.
      & Dr. Andrew Jones asserts that VO$_2$ max and lactate thresholds are critical to performance. & NEI & True \\
      \midrule

      Examples like Kipchoge's sub-2 marathon (2019, Vienna) demonstrate how overriding the central governor (with drafting, optimized shoes, and laser pacing) can push human limits.
      & Kipchoge's sub-2 marathon in 2019, in Vienna, involved strategies to override the central governor. & NEI & True \\
      \midrule

      The Central Governor Model (CGM) is a theoretical framework that attempts to explain how the brain regulates exercise intensity and endurance performance.
      & The Central Governor Model (CGM) is a theoretical framework that explains how the brain regulates exercise intensity and endurance performance. & support & True \\
      \midrule

      It posits that the central nervous system (CNS) plays a crucial role in controlling exercise intensity based on perceived energy availability and homeostatic needs.
      & The Central Governor Model posits that the central nervous system (CNS) controls exercise intensity based on perceived energy availability and homeostatic needs. & support & True \\
      \midrule

      In the 2018 Tour de France, Geraint Thomas used mental strategies to manage fatigue during the final stages of the race.
      & Geraint Thomas used mental strategies to manage fatigue during the 2018 Tour de France. & support & True \\
      \bottomrule
    \end{tabularx}
    \caption{Sentence-level Claim Verification Results.}
    \label{tab:sent_claim_label_agree}
    \vspace{-6pt}
\end{table*}

\subsection{Details of Metrics}
\label{appendix:details_metrics}
Based on the consideration of fine-grained indicators, we use a series of more detailed evaluation metrics in the process of evaluating fact-checking frameworks.

\paragraph{Sentence-level alignment and coverage (Cov).}
The common scored unit is a ground-truth source sentence. Sentence-level outputs are aligned directly. For claim- or subclaim-level systems, we map each output to its highest-similarity source sentence with \verb|bge-m3| and retain mappings with cosine similarity at least 0.85. If multiple verdicts map to one sentence, we aggregate them using the priority \textsc{Refute}, \textsc{NEI}, then \textsc{Support}; internal conflicting-evidence outputs are mapped to \textsc{NEI}. Coverage is the proportion of ground-truth source sentences that receive at least one aligned verdict. The matched-coverage analysis in Appendix~\ref{app:matched-coverage} further restricts all compared systems to an identical shared sentence subset.

\paragraph{Details of $F_1@K$}
Following the formulation in \textsc{VeriScore}, we report $F_1@K$ to provide a balanced assessment of both verification accuracy and information coverage. 
While standard precision measures the correctness of generated outputs, it fails to penalize overly reticent models that generate very few claims. 
$F_1@K$ addresses this by introducing a fixed ideal volume $K$ into the recall calculation. 
Formally, it is the harmonic mean of Precision and $\text{Recall}@K$:
\begin{equation}
    F_1@K = \frac{2 \cdot \text{Prec} \cdot \text{Rec}@K}{\text{Prec} + \text{Rec}@K}
\end{equation}
where $\text{Prec} = |S_{\text{sup}}| / |S_{\text{total}}|$ is the proportion of support units among all generated units, and $\text{Rec}@K = |S_{\text{sup}}| / K$ measures the volume of support information relative to the expected standard $K$.
The definition of $K$ varies by method to ensure fair comparison. 
For \textsc{Safe} and \textsc{VeriScore}, which operate at the atomic level, $K$ represents the average number of atomic claims in the human-annotated ground truth (or the dataset average). 
In contrast, for \textsc{ElementCheck}, we calibrate $K$ to the sentence level: here, $K$ denotes the average number of verifiable sentences per response in the dataset. 
This adaptation reflects our framework's emphasis on sentence-level verification elements rather than decomposed atomic facts.

\paragraph{Soft-Semantic F1.} 
To quantify the structural consistency between element graphs generated by different backbones, we employ the \textbf{Soft-Semantic $F_1$} metric. Let $\mathcal{G}_A$ and $\mathcal{G}_B$ denote the element graphs generated by model $A$ and model $B$. We define $\mathcal{S}_A$ and $\mathcal{S}_B$ as the corresponding sets of graph elements derived from these graphs, where $\mathcal{S}$ refers to either the set of unique entity nodes or the set of linearized relation edges. To account for semantic variations rather than strict lexical matching, we utilize the \verb|BGE-M3| encoder to map each element $x \in \mathcal{S}$ into a normalized semantic vector $\mathbf{v}_x$.We calculate the bidirectional consistency by measuring how well the semantic content of one set is covered by the other. Specifically, we define \textit{Semantic Recall} ($R_{sem}$) and \textit{Semantic Precision} ($P_{sem}$) as the average maximum cosine similarity:
\begin{equation}
\begin{aligned}
    R_{sem} &= \frac{1}{|\mathcal{S}_A|} \sum_{x \in \mathcal{S}_A} \max_{y \in \mathcal{S}_B} (\mathbf{v}_x^\top \mathbf{v}_y), \\
P_{sem} &= \frac{1}{|\mathcal{S}_B|} \sum_{y \in \mathcal{S}_B} \max_{x \in \mathcal{S}_A} (\mathbf{v}_y^\top \mathbf{v}_x).
\end{aligned}
\end{equation}
The final stability score is computed as the harmonic mean of these two directional coverage scores: 
\begin{equation}
    F_1 = \frac{2 \cdot (P_{sem} \cdot R_{sem})}{(P_{sem} + R_{sem})},
\end{equation}
For structural element connecting a head $h$ and a tail $t$, we linearize them into the sequence $s = \texttt{"[CLS] } h \texttt{ [SEP] } t \texttt{"}$ prior to embedding.

\section{Extended Results}

\subsection{Coverage-Controlled Statistical Analysis}
\label{app:matched-coverage}
Table~\ref{tab:matched-coverage} reports the point estimates on the 2,468 source sentences jointly covered by ElementCheck, VeriScore, and FastFact. We additionally apply one-sided paired bootstrap tests over this shared evaluation set and Holm correction across the four planned comparisons in Table~\ref{tab:matched-significance}.

\begin{table}[ht]
    \centering
    \small
    \resizebox{\columnwidth}{!}{%
    \begin{tabular}{lrrr}
      \toprule
      Comparison & $\Delta$ BAcc & Holm $p$ & 95\% CI \\
      \midrule
      VeriScore, BAcc-3 & +0.05372 & 0.00120 & [0.0214, 0.0868] \\
      FastFact, BAcc-3 & +0.09176 & 0.000400 & [0.0524, 0.1295] \\
      FastFact, BAcc-2 & +0.08655 & 0.000400 & [0.0534, 0.1188] \\
      VeriScore, BAcc-2 & $-$0.00591 & 0.6615 & [$-$0.0324, 0.0202] \\
      \bottomrule
    \end{tabular}
    }
    \caption{Paired comparisons (ElementCheck minus baseline) on the shared sentence set.}
    \label{tab:matched-significance}
    \vspace{-6pt}
\end{table}

The three-class gains over both baselines and the binary gain over FastFact remain significant; the binary difference from VeriScore does not. Restricting coverage removes differences in which sentences are scored, but does not make the underlying retrieval, extraction, and verification pipelines identical. We therefore use this analysis to separate coverage from matched-set accuracy, not to attribute the remaining difference to one component. The archived rebuttal record does not preserve the bootstrap repetition count or random seed; we report the archived statistics without reconstructing those metadata.

\subsection{Matched-Budget Router Controls}
\label{app:router-control}
For each FastFact-Sent sentence, we precompute a GPT-4o-mini direct verdict and a flat-element verdict using the same sentence and evidence. A router selects which verdict becomes the final prediction. Every router sends exactly 37.13\% of sentences to the flat-element path, matching ElementCheck's routing fraction; routers select the same number, but not necessarily the same sentences. The graph rule uses connectivity and diameter. The surface baselines rank by entity count or sentence length, and the random baseline samples the same budget.

For LLM routing, GPT-4o-mini and GPT-5 receive the same sentence-only complexity prompt (Figure~\ref{fig:routing-gate}) and output a score in $[0,1]$; the top-scoring sentences fill the fixed budget. Table~\ref{tab:matched-router} counts only incremental gate tokens. Consequently, zero added tokens for graph topology means that its deterministic decision is free after the element graph has already been extracted, not that graph extraction itself is free. The archived rebuttal record does not preserve heuristic tie-breaking or random-router tie-breaking, seed, and aggregation metadata; the small router differences are therefore interpreted descriptively.

\subsection{Reliability and Open-Verifier Compatibility}
\label{app:reliability}
We reran the ElementCheck verification stage three times while fixing the input sentences. Mean \mbox{BAcc-2} is 0.7150 (SD 0.0018) and mean \mbox{BAcc-3} is 0.5330 (SD 0.0037); the original run obtains 0.7131 and 0.5359, respectively.

Because the perturbation analysis identifies spurious pairs as the most harmful error, we conduct a precision-focused audit of 400 pairs sampled from the 18,310 extracted pairs, split evenly between graph-simple and graph-complex cases. Two independent judges from model families different from the GPT-4o-mini extractor label whether each pair is explicitly grounded in its source sentence. Gemini-2.5-Flash estimates overall precision at 0.895 (95\% CI [0.861, 0.921]), with 0.920 on simple and 0.870 on complex cases. Claude-Sonnet-4.5 estimates 0.875 (95\% CI [0.839, 0.904]), with 0.875 and 0.874, respectively. Raw agreement between the two judges is 91.7\%. The principal error is pairing entities that co-occur without an explicitly stated factual connection; this audit estimates precision rather than extraction recall.

We also replace only the verifier component with open models, while keeping extraction and retrieval fixed to GPT-4o-mini outputs and using non-reasoning inference. GPT-4o-mini obtains BAcc-2/BAcc-3 of 0.7131/0.5359, Qwen3.6-27B obtains 0.7512/0.5843, and Qwen3.5-35B-A3B obtains 0.7465/0.5762. These results demonstrate component-level compatibility with the tested open verifiers; they are not end-to-end open-pipeline results.

\subsection{Retrieval-Controlled Comparison}
\label{app:retrieval-control}
The main systems retain their released retrieval procedures. ElementCheck uses Serper top-10 snippets, whereas the original FastFact run uses Jina documents. Under a same-configuration ElementCheck replication, BAcc-2/BAcc-3 changes from 0.7131/0.5359 to 0.7209/0.5376. Replacing FastFact's Jina backend with Serper top-10 while fixing its remaining pipeline changes 0.6430/0.4684 to 0.6503/0.4847. The remaining gap under Serper is descriptive rather than causal because query construction, extraction, representation, verifier behavior, and aggregation still differ.

\subsection{Cost and Efficiency Across Backbones}
\label{app:overhead}

Table~\ref{tab:overhead-backbone} reports cost and efficiency across additional backbones. The relative ordering is consistent with the main experiments: VeriScore is generally more expensive, FastFact is lightweight but verifies fewer units, and ElementCheck keeps a moderate cost while preserving sentence-level verification coverage.

\begin{table}[ht]
    \centering
    \small
    \resizebox{\columnwidth}{!}{%
    \begin{tabular}{llcccc}
      \toprule
      Backbone & Method & $\text{Avg}_\text{Calls}$ & $\text{Avg}_\text{Claims}$ & Input (K) & Output (K)\\
     \midrule
      \multirow{3}{*}{GPT-5.1}
      & VeriScore & 70.83 & 52.51 & 46.02 & 7.15\\
      & FastFact & 36.77 & 6.33 & 29.09 & 2.15\\
      & ElementCheck & 33.37 & 25.09 & 30.13 & 5.40\\
     \midrule
      \multirow{3}{*}{Qwen3-235B}
      & VeriScore & 64.97 & 45.12 & 41.64 & 6.08\\
      & FastFact & 32.72 & 7.24 & 23.31 & 2.08\\
      & ElementCheck & 37.58 & 25.41 & 30.49 & 6.23\\
      \bottomrule
    \end{tabular}
    }
    \caption{Cost and efficiency across backbones on FastFact-Sent. $\text{Avg}_\text{Calls}$: average API calls per response; $\text{Avg}_\text{Claims}$: average verified units; Input/Output: total tokens (K).}
    \label{tab:overhead-backbone}
    \vspace{-6pt}
\end{table}

\subsection{Fine-Grained 3-Class Results}
\label{app:3class-results}

Table~\ref{tab:3class} reports the 3-class setting, where Support, Refute, and NEI are evaluated separately. This setting is stricter than the binary Support and NoSupport evaluation because the model must distinguish unsupported claims from evidence-insufficient claims. ElementCheck remains competitive across backbones and domains, indicating that the element-guided verification process improves not only binary factuality detection but also finer-grained label assignment.

\begin{table}[t]
  \vspace{-4pt}
  \centering
  \begin{small}
  \resizebox{0.48\textwidth}{!}{%
    \begin{tabular}{l*{6}{c}}
      \toprule
       Method & ExpertQA & FCBench & Bio & HelloBench & LongFact & Avg\\
      \midrule
      \rowcolor{gray!15}
      \multicolumn{7}{c}{\textbf{\texttt{GPT-4o-mini}}} \\
      \midrule
      VeriScore & 44.5 & 49.1 & 51.9 & 48.5 & 52.1 & 49.2 \\
      FastFact & 47.4 & 42.6 & 55.2 & 44.4 & 41.7 & 46.3 \\
      ElementCheck & 42.3 & \textbf{52.2} & \textbf{60.2} & \textbf{49.2} & 51.7 & \textbf{51.1} \\
      \midrule
      \rowcolor{gray!15}
      \multicolumn{7}{c}{\textbf{\texttt{GPT-5.1}}} \\
      \midrule
      VeriScore & 46.1 & 43.9 & 54.8 & 51.2 & 49.6 & 49.1 \\
      FastFact & \textbf{52.4} & 45.4 & 50.6 & 52.4 & \textbf{55.8} & 51.3 \\
      ElementCheck & 50.8 & \textbf{51.3} & \textbf{53.9} & \textbf{53.9} & 51.8 & \textbf{54.5} \\
      \midrule
      \rowcolor{gray!15}
      \multicolumn{7}{c}{\textbf{\texttt{DeepSeek-V3.2}}} \\
      \midrule
      VeriScore & 39.8 & 38.5 & 41.9 & 43.1 & 45.8 & 41.8 \\
      FastFact & 33.3 & 47.3 & 44.1 & 38.1 & 42.2 & 41.0 \\
      ElementCheck & \textbf{40.7} & \textbf{52.1} & \textbf{57.3} & \textbf{49.3} & \textbf{51.8} & \textbf{50.3} \\
      \midrule
      \rowcolor{gray!15}
      \multicolumn{7}{c}{\textbf{\texttt{Qwen3-235B-A22B-Instruct-2507}}} \\
      \midrule
      VeriScore & \textbf{48.8} & 43.2 & 48.0 & 43.5 & 48.8 & 46.5 \\
      FastFact & 49.4 & 40.9 & 52.5 & 51.5 & 52.2 & 49.3 \\
      ElementCheck & 48.7 & \textbf{55.2} & \textbf{62.4} & \textbf{56.4} & \textbf{58.6} & \textbf{56.2} \\
      \bottomrule
    \end{tabular}
    }
  \end{small}
    \caption{3-class balanced accuracy (BAcc-3: Support, Refute, and NEI) on FastFact-Sent across four backbones. ElementCheck achieves the best average score in each backbone group under the more challenging fine-grained evaluation.}
    \label{tab:3class}
  \vspace{-6pt}
\end{table}

\subsection{Backbone-Level \texorpdfstring{$F_1@K$}{F1@K} Analysis}
\label{app:f1k}

\begin{figure}[ht]
  \centering
  \includegraphics[width=\columnwidth]{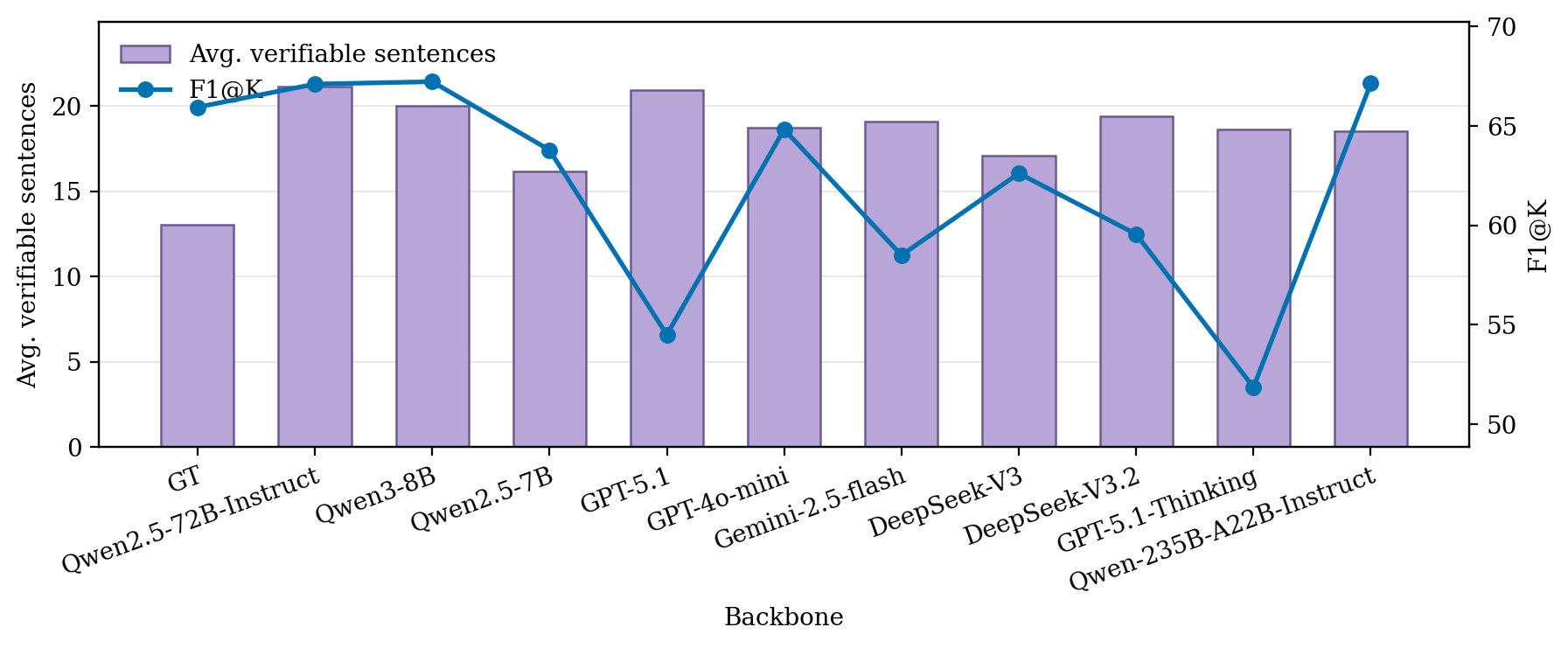}
  \caption{Backbone-level comparison between average verifiable sentence count and $F_1@K$.}
  \label{fig:different-f1}
  \vspace{-8pt}
\end{figure}

Figure~\ref{fig:different-f1} compares the average number of verifiable sentences with $F_1@K$ across backbone models. The figure shows that generating or identifying more verifiable sentences does not by itself guarantee stronger factuality evaluation. Backbones differ in how well they balance coverage with verification precision, which motivates reporting both coverage-sensitive metrics and class-balanced accuracy in the main experiments.

\subsection{Claim-Refinement Baselines}
\label{app:refinement}

Table~\ref{tab:refinement} compares ElementCheck with two claim-refinement baselines using the same verifiable sentences from our FastFact-Sent preprocessing. The comparison shows that explicit claim refinement can improve some fine-grained decisions, but it introduces additional decomposition and refinement steps. ElementCheck remains close in performance while using a simpler sentence-level pipeline.

\begin{table}[ht]
    \centering
    \small
    \resizebox{\columnwidth}{!}{%
    \begin{tabular}{lcccccc}
      \toprule
      \multirow{2}{*}{Dataset} & \multicolumn{2}{c}{ElementCheck} & \multicolumn{2}{c}{DnDScore} & \multicolumn{2}{c}{VeriFact}\\
      & BAcc-3 & BAcc-2 & BAcc-3 & BAcc-2 & BAcc-3 & BAcc-2\\
     \midrule
      ExpertQA & 42.3 & 67.2 & \textbf{50.5} & 67.2 & 46.8 & 66.5\\
      FCBench & \textbf{52.2} & 64.0 & 51.1 & 65.7 & 52.3 & \textbf{69.7}\\
      Bio & \textbf{60.2} & 74.3 & 58.5 & 72.2 & 59.6 & \textbf{78.1}\\
      HelloBench & 49.2 & \textbf{69.0} & \textbf{56.8} & 68.3 & 48.3 & 64.5\\
      LongFact & 51.7 & 69.7 & \textbf{53.7} & \textbf{70.4} & 50.2 & 69.1\\
     \midrule
      \textbf{Avg} & 51.1 & 68.8 & \textbf{54.1} & 68.7 & 51.5 & \textbf{69.6}\\
      \bottomrule
    \end{tabular}
    }
    \caption{Comparison with claim-refinement baselines on FastFact-Sent with GPT-4o-mini. BAcc-3 and BAcc-2 are reported per dataset.}
    \label{tab:refinement}
    \vspace{-6pt}
\end{table}

\subsection{Per-Class Error Analysis}
\label{sec:error-analysis}

To understand where ElementCheck's gains originate, we compare per-class accuracy between Direct Verification and ElementCheck under GPT-4o-mini and GPT-5.1 (Table~\ref{tab:perclass}).

\begin{table}[ht]
    \centering
    \small
    \resizebox{\columnwidth}{!}{%
    \begin{tabular}{lcccccc}
      \toprule
      \multirow{2}{*}{Model} & \multicolumn{2}{c}{BAcc} & \multicolumn{2}{c}{Refute Acc} & \multicolumn{2}{c}{NEI Acc}\\
      & Direct & EC & Direct & EC & Direct & EC\\
     \midrule
         GPT-4o-mini & 64.7 & 68.8 & 14.6 & 25.3 & 38.6 & 48.1\\
         GPT-5.1 & 64.4 & 71.3 & 19.9 & 25.2 & 66.2 & 81.0\\
       \bottomrule
    \end{tabular}
    }
    \caption{Per-class accuracy comparison between Direct Verification and ElementCheck. Gains concentrate on Refute and NEI. EC indicates ElementCheck.}
    \label{tab:perclass}
    \vspace{-6pt}
\end{table}

\paragraph{Gains concentrate on hard classes.} ElementCheck's improvement primarily comes from Refute and NEI rather than Support. This is consistent with element-based verification being most useful when the verifier must identify contradictions or evidence gaps.

\paragraph{Error patterns differ by backbone.} GPT-4o-mini tends to over-predict Support, while GPT-5.1 is more conservative and assigns NEI more often. ElementCheck's remaining errors are predominantly gold-Support sentences shifted to NEI, indicating a more conservative error pattern on borderline claims.

\paragraph{When does ElementCheck not help?} ElementCheck provides limited additional value for straightforward factual assertions that are already directly supported by retrieved evidence. Its gains concentrate on semantically complex cases involving negation, qualification, or multi-entity relations. On weaker backbones, coverage can also be limited by the Verifiable Sentence Acquisition stage, suggesting that upstream sentence acquisition remains an important factor for future improvement.

Scalability analysis across document lengths (Appendix~\ref{app:scalability}) further shows that graph construction remains controlled as responses become longer.

\section{Robustness and Stability Analysis}
\label{app:robust-stability}

\subsection{Cross-Backbone Graph Stability}
\label{app:stability}

\begin{figure}[ht]
  \centering
  \includegraphics[width=\columnwidth]{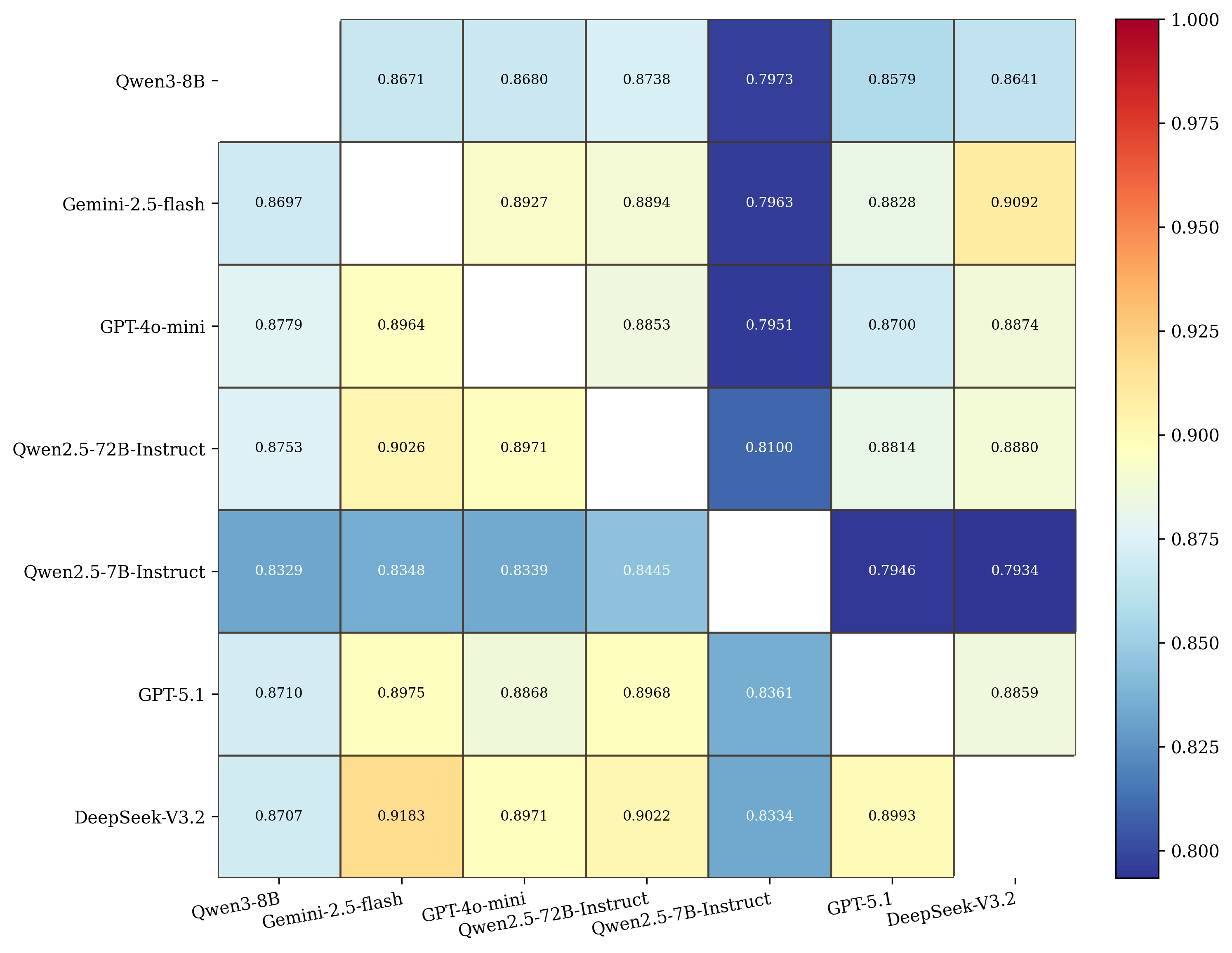}
  \caption{Element graph similarity across backbone models. The upper triangle reports entity-node consistency, and the lower triangle reports relation consistency.}
  \label{fig:graph_analysis}
  \vspace{-8pt}
\end{figure}

ElementCheck uses the element graph as a structural interface between sentence acquisition and verification. We therefore evaluate its similarity when extracted by different backbone models. Figure~\ref{fig:graph_analysis} reports pairwise Soft-Semantic F1 scores over entity nodes and relations. The results show similar extracted content across the evaluated backbones, without establishing extractor invariance.

To further separate extraction stability from verification ability, Table~\ref{tab:crossmodel} decouples the graph extractor and verifier. When the verifier is fixed, changing the graph extractor produces only small performance variation. This suggests that the main source of performance difference is the verifier's reasoning capacity rather than instability in the extracted graph.

\begin{table}[ht]
    \centering
    \small
    \begin{tabular}{llcc}
      \toprule
      Graph & Verifier & BAcc-2 & BAcc-3\\
     \midrule
         GPT-4o-mini & GPT-5.1 & 72.04 & 53.33\\
         GPT-4o-mini & Qwen3-235B & 74.03 & 54.47\\
         GPT-5.1 & GPT-4o-mini & 71.13 & 50.79\\
         GPT-5.1 & Qwen3-235B & 74.56 & 55.90\\
         Qwen3-235B & GPT-4o-mini & 72.05 & 52.70\\
         Qwen3-235B & GPT-5.1 & 73.81 & 51.73\\
       \bottomrule
    \end{tabular}
    \caption{Cross-model stability with graph extraction and verification decoupled. Under a fixed verifier, performance changes less across graph extractors than across verifier backbones.}
    \label{tab:crossmodel}
    \vspace{-6pt}
\end{table}

\subsection{Perturbation Study}
\label{app:perturbation}

To measure how downstream verification changes under noisy upstream element extraction, we inject synthetic noise into the extracted element pairs before verification. Dropout simulates missed extraction, Swap tests sensitivity to corrupted entities, Addition simulates spurious pairs, and Shuffle tests whether pair direction alone drives decisions. All experiments use GPT-4o-mini on FastFact-Sent.

\begin{table}[ht]
    \centering
    \small
    \resizebox{\columnwidth}{!}{%
    \begin{tabular}{llcccc}
      \toprule
      Type & $p$ & BAcc-3 & $\Delta$ BAcc-3 & BAcc-2 & $\Delta$ BAcc-2\\
     \midrule
      \textbf{Original} & --- & 0.536 & --- & 0.713 & ---\\
     \midrule
      \multirow{4}{*}{Dropout}
      & 0.1 & 0.531 & $-$0.005 & 0.707 & $-$0.007\\
      & 0.2 & 0.534 & $-$0.002 & 0.712 & $-$0.002\\
      & 0.3 & 0.522 & $-$0.014 & 0.706 & $-$0.008\\
      & 0.5 & 0.529 & $-$0.008 & 0.712 & $-$0.001\\
     \midrule
      \multirow{4}{*}{Shuffle}
      & 0.1 & 0.527 & $-$0.009 & 0.709 & $-$0.004\\
      & 0.2 & 0.526 & $-$0.010 & 0.711 & $-$0.003\\
      & 0.3 & 0.525 & $-$0.011 & 0.709 & $-$0.004\\
      & 0.5 & 0.525 & $-$0.011 & 0.705 & $-$0.009\\
     \midrule
      \multirow{4}{*}{Swap}
      & 0.1 & 0.527 & $-$0.010 & 0.703 & $-$0.010\\
      & 0.2 & 0.509 & $-$0.027 & 0.700 & $-$0.014\\
      & 0.3 & 0.515 & $-$0.021 & 0.705 & $-$0.009\\
      & 0.5 & 0.495 & $-$0.041 & 0.681 & $-$0.032\\
     \midrule
      \multirow{4}{*}{Addition}
      & 0.1 & 0.445 & $-$0.092 & 0.611 & $-$0.102\\
      & 0.2 & 0.433 & $-$0.104 & 0.609 & $-$0.105\\
      & 0.3 & 0.457 & $-$0.079 & 0.638 & $-$0.076\\
      & 0.5 & 0.455 & $-$0.081 & 0.636 & $-$0.078\\
      \bottomrule
    \end{tabular}
    }
    \caption{Verification performance under controlled perturbation. Original: no perturbation.}
    \label{tab:perturbation}
    \vspace{-6pt}
\end{table}

The results reveal a clear pattern: ElementCheck is relatively robust to pair direction reversal and omission, moderately sensitive to entity replacement, and most sensitive to spurious pair addition. This indicates that extraction \textbf{precision matters more than recall} for stable downstream verification: incomplete anchors can often be recovered from context, whereas false structural cues may misdirect the verifier.

\subsection{Alternative Representations}
\label{app:representations}

We compare ElementCheck's element graph representation against four alternatives under a unified evaluation setting. All methods share the same upstream pipeline; only the representation and verification granularity differ.

\begin{table}[ht]
    \centering
    \small
    \resizebox{\columnwidth}{!}{%
    \begin{tabular}{lccccc}
      \toprule
      Method & BAcc-3 & BAcc-2 & Sup-Acc & Ref-Acc & NEI-Acc\\
     \midrule
      Direct & 0.463 & 0.647 & 0.857 & 0.146 & 0.387\\
      Triples (S,R,O) & 0.529 & 0.679 & 0.609 & 0.286 & 0.694\\
      SRL Frames & \textbf{0.543} & \textbf{0.718} & 0.751 & \textbf{0.289} & \textbf{0.588}\\
      Element Pairs (no graph) & 0.436 & 0.644 & 0.577 & 0.017 & 0.714\\
      \textbf{Element Graph (ours)} & 0.536 & 0.713 & \textbf{0.874} & 0.253 & 0.482\\
      \bottomrule
    \end{tabular}
    }
    \caption{Comparison of structured representations with the upstream pipeline fixed. Element Graph approaches SRL balanced accuracy with lower token cost and the highest Support accuracy.}
    \label{tab:representations}
    \vspace{-6pt}
\end{table}

\paragraph{Token Efficiency.} Table~\ref{tab:token-cost} reports the per-method token cost. Element Graph provides an accuracy--cost trade-off among structured representations: it improves verification over direct sentence checking while avoiding the larger overhead introduced by SRL frames and triplet extraction.

\begin{table}[ht]
    \centering
    \small
    \begin{tabular}{lcc}
      \toprule
      Method & Sent. Tokens & Relative Cost\\
     \midrule
      Direct & 695 & 1.00$\times$\\
      \textbf{Element Graph (ours)} & 1,123 & \textbf{1.62$\times$}\\
      Element Pairs (no graph) & 1,158 & 1.67$\times$\\
      SRL Frames & 1,446 & 2.08$\times$\\
      Triples & 2,001 & 2.88$\times$\\
      \bottomrule
    \end{tabular}
    \caption{Token cost comparison of structured representations.}
    \label{tab:token-cost}
    \vspace{-6pt}
\end{table}

\paragraph{Key Findings.} SRL gives slightly higher BAcc-2 and BAcc-3, whereas Element Graph reaches comparable balanced accuracy at lower token cost and gives the highest Support accuracy. Triplet extraction is more expensive, and the flat Element Pairs variant is substantially less accurate. The comparison therefore supports a granularity--context--efficiency trade-off rather than universal superiority of one representation.

\subsection{Scalability Analysis}
\label{app:scalability}

We analyze how graph complexity scales with document length on FastFact-Sent. Scalability is considered at two levels: graph construction as document length increases, and verification cost after graph-guided routing.

\begin{table}[ht]
    \centering
    \small
    \resizebox{\columnwidth}{!}{%
    \begin{tabular}{lccccc}
      \toprule
      Length Bin & Avg Words & Avg Pairs & Components & Pairs/Sent & Direct Ratio\\
     \midrule
      $<$150 & 98.0 & 9.5 & 1.2 & 1.86 & 0.426\\
      150--300 & 226.4 & 30.2 & 3.2 & 2.52 & 0.554\\
      300--500 & 403.9 & 47.1 & 7.2 & 1.76 & 0.616\\
      500--800 & 630.5 & 86.4 & 11.0 & 2.00 & 0.687\\
      800+ & 987.8 & 142.1 & 14.7 & 2.07 & 0.685\\
      \bottomrule
    \end{tabular}
    }
    \caption{Complexity growth by document length on FastFact-Sent. Pairs/Sent: average element pairs per sentence; Direct Ratio: proportion of sentences routed to direct verification.}
    \label{tab:scalability}
    \vspace{-6pt}
\end{table}

\paragraph{Key Findings.} Graph size grows with document length, but the growth is well structured rather than monolithic. Longer responses naturally split into more graph components, and the per-sentence number of element pairs remains bounded. This supports the scalability of graph-guided routing: longer responses do not necessarily require proportionally more expensive verification, because many sentences remain structurally simple and can be handled by the direct path.

\subsection{Graph Statistics}
\label{app:graph-stats}

Table~\ref{tab:graph-dist} reports the element graph diameter distribution on FastFact-Sent. Most connected graphs have small diameters, while disconnected graphs reflect sentences with multiple loosely coupled semantic units. This distribution motivates $\delta=3$ as a practical operating point for the evaluated setting.

\begin{table}[ht]
    \centering
    \small
    \begin{tabular}{lcc}
      \toprule
      Category & Count & Ratio\\
     \midrule
      Disconnected & 2,010 & 29.1\%\\
      $d=1$ & 324 & 4.7\%\\
      $d=2$ & 2,712 & 39.2\%\\
      $d=3$ & 1,310 & 18.9\%\\
      $d=4$ & 434 & 6.3\%\\
      $d=5$ & 101 & 1.5\%\\
      $d \geq 6$ & 23 & 0.3\%\\
      \bottomrule
    \end{tabular}
    \caption{Element graph diameter distribution on FastFact-Sent (GPT-4o-mini).}
    \label{tab:graph-dist}
    \vspace{-6pt}
\end{table}

\section{Prompt}


The prompts of Verifiable Sentence Acquisition, Element Extraction, Element-based Verification, Claim Re-exaction and Direct Verification correspond to Figure~\ref{fig:sentence acquisition}, Figure~\ref{fig:element extraction}, Figure~\ref{fig:element-base verification}, Figure~\ref{fig:claim-re-exaction} and Figure~\ref{fig:direct verification}.

\begin{AIbox}{Prompt of Verifiable Sentence Acquisition.}
{\color{black}\bf \large User Prompt:}
\tcblower
\vspace{1mm}

You are a preprocessing assistant for factuality checking.
Given a question and a paragraph in the response that is divided into a sentence list.

Your task is to perform coreference resolution and verifiable verification on each sentence in the list based on the question, and return the processed sentences and a list.

\textbf{Strictly follow the process below:}

1. Verifiable Check: Return an empty string for sentences that cannot be verified.

A sentence is verifiable if it makes an objectiv statement that can in principle be verifiable; 

A sentence consisting entirely of subjective statements, imperative sentences, interrogative sentences, or transitional sentences connecting contexts is unverifiable.

If a sentence mixes subjective and objective elements, but contains at least one objective, verifiable claim as described above, remove the subjective elements and retain only the verifiable clauses.

2. Coreference Resolution: Use the question and all segments as context to resolve pronouns and vague references in each sentence, such as: he, she, it, they, this, that, these, those, the character, the story, some, others, etc.

3. Sentence Rewrite: 

For sentences that cannot be verified, return an empty string "". 

For sentences that can be verified, make only minimal changes: complete coreference resolution and remove some conjunctions or clauses without verification information. 

If the sentence does not require modification, return the original sentence.

Here are some example below:

\textbf{Example 1:}
\begin{lstlisting}[style=prompt]
Example 1:
Question:
What is the history behind the creation of the periodic table, and what impact did it have on modern science and chemistry?
Text List:
["The periodic table is a way of organizing the elements based on their atomic number, chemical properties and physical properties.", "It is a very useful tool for modern science and chemistry because it helps to predict how elements will react with each other, understand the trends in their properties and discover new elements."]
Output:
["The periodic table is a way of organizing the elements based on their atomic number, chemical properties and physical properties.", "The periodic table is a useful tool for modern science and chemistry because it helps scientists predict how elements will react with each other, understand trends in their properties, and discover new elements."]    
\end{lstlisting}

\textbf{Example 2:}
\begin{lstlisting}[style=prompt]
Question:
What inspired Stephen Hawking to pursue a career in theoretical physics?
Text List:
["Here are some examples of how Stephen Hawking inspired others to pursue careers in theoretical physics.", "According to some sources, Hawking was inspired by his father's interest in medicine and his mother's enthusiasm for philosophy and politics.", "He also had a natural curiosity and passion for science and mathematics from an early age.", "He was influenced by his teachers and mentors at Oxford and Cambridge, such as Robert Berman and Dennis Sciama ."]
Output:
["", "Stephen Hawking was inspired by his father's interest in medicine and his mother's enthusiasm for philosophy and politics.", "Stephen Hawking had a natural curiosity and passion for science and mathematics from an early age.", "Stephen Hawking was influenced by his teachers and mentors at Oxford and Cambridge, including Robert Berman and Dennis Sciama."]
\end{lstlisting}

\textbf{Example 3:}
\begin{lstlisting}[style=prompt]
Question:
What is one of the most significant climate science findings in recent years?
Text List:
["One major climate-science finding in recent years is the sharply accelerating ice-sheet mass loss in Antarctica.","According to satellite gravimetry analyses from missions such as NASA's GRACE and GRACE-FO, Antarctica has been losing ice several times faster than in the 1990s, with West Antarctica showing patterns that some researchers describe as tipping-like.","Scientists attribute this rapid mass loss to several interacting physical drivers:"]

Output:
["One major climate-science finding in recent years is the sharply accelerating ice-sheet mass loss in Antarctica.", "Antarctica has been losing ice several times faster than in the 1990s, with West Antarctica showing patterns that some researchers describe as tipping-like.", ""]
\end{lstlisting}

{\color{black}\bf Input:}\\
Question: {\color{deepblue}\bf \{question\}} \\
Text List: {\color{deepblue}\bf \{text\_list\}} \\
{\color{black}\bf Output:}\\

\end{AIbox}
\captionof{figure}{Prompt of Verifiable Sentence Acquisition.}
\label{fig:sentence acquisition}

\begin{AIbox}{Prompt of Element Extraction.}
{\color{black}\bf \large User Prompt:}
\tcblower
\vspace{1mm}

You are an information extraction system that works sentence by sentence.

For each sentence in a sentence List:

1. Identify noun entities and drop leading determiners like "the/a/an", the entities should include:

  - proper names

  - numeric and time expressions 
  
  - Noun phrases in the sentence
  
2. Using only the sentence's relational descriptions, extract connected entity pairs in the form ["head", "tail"]. The relations include:
  
  - predicative relations
  
  - attributive relations
  
  - modifying relations
  
  - appositive or exemplification structures
  
  - temporal or spatial anchoring
  
  - causal dependencies

3. If a mention is a pronoun or a coreferent form (e.g., this/that/these/those/it/they/the xxx), you must resolve it using the current or previous sentence and record the resolution.

4. Skip sentences with non-verifiable or imperative structure (e.g., "Here are some examples.", "Do you have a favorite one?")

\textbf{Coreference}

\begin{itemize}
    \item Resolve within a 2-3 sentence window to the most specific antecedent.
    \item If uncertain, skip that pair.
    \item Do not treat relative pronouns (e.g., "which", "that", "who") in restrictive or non-restrictive clauses as true coreference mentions.
    \item Record each resolution as ["mention", "resolved\_to"]
\end{itemize}

\textbf{Output Format is JSON}

Return a JSON array where each element corresponds to one sentence:
\begin{lstlisting}[style=prompt]
[{{
    "pairs":[["head1", "tail1"], ["head2", "tail2"], ...]
    "coref":[["this", "entity1"]]
}}]   
\end{lstlisting}

\textbf{Example:}

\begin{lstlisting}[style=prompt]
Text:
'On June 11, 1770, James Cook\'s ship Endeavour ran aground on the Great Barrier Reef.', 'This accident on that day marked the first recorded European shipwreck on the reef.']
Output:
[
  {{
    "pairs": [
      ["James Cook's ship Endeavour", "Great Barrier Reef"],
      ["James Cook's ship Endeavour", "June 11, 1770"],
      ["shipwreck", "June 11, 1770"]
    ],
    "coref": []
  }},
  {{
    "pairs": [
      ["accident", "shipwreck"],
      ["accident", "June 11, 1770"],
      ["shipwreck", "Great Barrier Reef"]
    ],
    "coref": [
      ["This accident", "the shipwreck of James Cook's ship Endeavour on the Great Barrier Reef on June 11, 1770"],
      ["that day", "June 11, 1770"],
      ["the reef", "Great Barrier Reef"]
    ]
  }}
]

Text:
['The protein retention capability of solid reconstituted corneal cells (sRCCs) is superior to that of extruded RCCs (eRCCs), and this difference has significant implications for antigen presentation in vaccine development.', 'These retain their original dense, fibrous matrix structure, which provides a high surface area and physical stability.']

Output:
[
  {{
    "pairs": [
      ["protein retention capability", "solid reconstituted corneal cells (sRCCs)"],
      ["protein retention capability", "extruded RCCs (eRCCs)"],
      ["difference", "implications"],
      ["implications", "antigen presentation"],
      ["antigen presentation", "vaccine development"]
    ],
    "coref": [
      ["this difference", "difference in protein retention capability between sRCCs and eRCCs"]
    ]
  }},
  {{
    "pairs": [
      ["solid reconstituted corneal cells (sRCCs)", "matrix structure"],
      ["matrix structure", "high surface area"],
      ["matrix structure", "physical stability"]
    ],
    "coref": [
      ["These", "solid reconstituted corneal cells (sRCCs)"],
      ["their", "solid reconstituted corneal cells (sRCCs)"]
    ]
  }}
] 
\end{lstlisting}

{\color{black}\bf Input:}\\
Text: {\color{deepblue}\bf \{text\}} \\
{\color{black}\bf Output:}\\

\end{AIbox}
\vspace{-1em}
\captionof{figure}{Prompt of Element Extraction.}
\label{fig:element extraction}

\begin{AIbox}{Prompt of Element-based Verification.}
{\color{black}\bf \large User Prompt:}
\tcblower
\vspace{1mm}

You are a precision fact-checking assistant.

You will be given:
\begin{itemize}
    \item a Sentence,
    \item a list of Locator Pairs,
    \item a set of Evidence snippets.
\end{itemize}

\textbf{Locator Pairs:}

1. Each locator pair consists of two entities from the Sentence.

2. \textbf{Contextual Anchors}: These pairs are NOT just isolated keywords. They anchor specific \textbf{modifiers, attributes, or actions} connecting these two entities within the Sentence context.

\textbf{Your task:}

1. \textbf{Locate Context}: Use each Locator Pair to identify the specific relationship, modification, or predicate assertion made in the Sentence.
2. \textbf{Verify}: Check if the Evidence supports that specific relationship or modification.
3. \textbf{Classify}: Assign exactly one of the following four labels to every locator pair based only on the Evidence.

Classifications:

1. \textbf{Support}

The Evidence confirms the specific relationship or modification described in the Sentence.

Synonyms and semantic equivalents are accepted.

2. \textbf{Not Enough Evidence}

The Evidence is missing, irrelevant, or too vague regarding the specific relationship.
  
Use this when the Evidence mentions the entities but does not verify the specific modification or action asserted in the Sentence.

3. \textbf{Conflicting Evidence}

The Evidence snippets contain internal contradictions regarding this specific relationship.

4. \textbf{Refute}

The Evidence explicitly contradicts the specific relationship or modification described in the Sentence.

\textbf{Output format:}

Your entire output must be a single valid JSON object with exactly the following structure. You must categorize every provided locator pair into one of these four lists.
\begin{lstlisting}[style=prompt]
{{
  "Support": [
    ["head_a", "tail_a"]
  ],
  "Not Enough Evidence": [
    ["head_b", "tail_b"]
  ],
  "Conflicting Evidence": [
    ["head_c", "tail_c"]
  ],
  "Refute": [
    ["head_d", "tail_d"]
  ]
}}
\end{lstlisting}

\textbf{Format constraints:}

1. The only keys you may output are `"Support"`, `"Not Enough Evidence"`, `"Conflicting Evidence"`, and `"Refute"`.

2. The value of each key must be a JSON array of locator pairs.

3. You must copy locator pairs verbatim from the given Locator Pairs list.

4. Each locator pair from the input must appear in exactly one of the four arrays.

5. If a category is empty, verify output `[]`.

6. Do NOT output any natural language explanation, comments, or headings.

{\color{black}\bf Input:}\\
Sentence: {\color{deepblue}\bf \{sentence\}} \\
Locator Pairs: {\color{deepblue}\bf \{pairs\}} \\
Evidence: {\color{deepblue}\bf \{evidence\}} \\
{\color{black}\bf Output:}\\

\end{AIbox}
\vspace{-1em}
\captionof{figure}{Prompt of Element Based Verification.}
\label{fig:element-base verification}

\begin{AIbox}{Prompt of Claim Re-exaction.}
{\color{black}\bf \large User Prompt:}
\tcblower
\vspace{1mm}

You are a fact-checking assistant. Your task is to generate ONE verifiable claim based on unverified elements from a sentence.

You will receive:
1. \textbf{Sentence}: The original sentence to be verified

2. \textbf{Unverified Elements}: A list of element pairs that could not be verified (marked as "not enough evidence")

\textbf{Task}

Based on the unverified elements, generate ONE specific, atomic, and independently verifiable claim that covers the most important unverified information.

\textbf{Requirements}

The claim should be:

- Atomic: Contains only one verifiable fact

- Self-contained: Can be understood without additional context

- Specific: Includes concrete entities, numbers, dates, or facts

- Searchable: Can be verified through web search

- Comprehensive: Covers the most critical unverified element(s)

If multiple elements are unverified, prioritize the most important or specific one.

\textbf{Output Format}

Return ONLY the claim as a single string (no quotes, no JSON array, just the claim text).

If no verifiable claim can be extracted, return an empty string: ""

\textbf{Examples}
\begin{lstlisting}[style=prompt]
Example 1:
Sentence: "Albert Einstein was born in Ulm, Germany in 1879 and won the Nobel Prize in 1921."
Unverified Elements: [["Albert Einstein", "Ulm"], ["Albert Einstein", "1879"]]
Output: Albert Einstein was born in Ulm, Germany in 1879.

Example 2:
Sentence: "The company reported revenue of $50 billion in 2023, a 15% increase from the previous year."
Unverified Elements: [["company", "$50 billion"], ["revenue increase", "15%"]]
Output: The company reported revenue of $50 billion in 2023, representing a 15% increase from the previous year.

Example 3:
Sentence: "She is a famous actress."
Unverified Elements: []
Output:
\end{lstlisting}

{\color{black}\bf Input:}\\
Sentence: {\color{deepblue}\bf \{sentence\}} \\
Unverified Elements: {\color{deepblue}\bf \{unverified\_elements\}} \\
{\color{black}\bf Output:}\\

\end{AIbox}
\vspace{-1em}
\captionof{figure}{Prompt of Claim Re-exaction.}
\label{fig:claim-re-exaction}

\begin{AIbox}{Prompt of Direct Verification.}
{\color{black}\bf \large User Prompt:}
\tcblower
\vspace{1mm}

You need to judge whether a claim is supported, refuted, or whether there is not enough evidence based on Google search results.

Below are the definitions of the three categories:

\textbf{support}: A claim is supported by the search results if everything in the claim is supported and nothing is contradicted by the search results. There can be some search results that are not fully related to the claim.

\textbf{refute}: A claim is refuted by the search results if something in the claim is contradicted by some search results. There should be no search result that supports the same part.

\textbf{not enough evidence}: A claim is inconclusive based on the search results if:

    - a part of a claim cannot be verified by the search results,

    - a part of a claim is supported and contradicted by different pieces of evidence.

You must output ONLY a \textbf{JSON} object in this exact format:
\begin{lstlisting}[style=prompt]
{{
    "label": "<support|refute|not enough evidence>"
}}
\end{lstlisting}

{\color{black}\bf Input:}\\
Claim: {\color{deepblue}\bf \{claim\}} \\
Evidence: {\color{deepblue}\bf \{evidence\}} \\
{\color{black}\bf Output:}\\

\end{AIbox}
\vspace{-1em}
\captionof{figure}{Prompt of Direct Verification.}
\label{fig:direct verification}

\begin{AIbox}{Prompt of Discriminator.}
{\color{black}\bf \large User Prompt:}
\tcblower
\vspace{1mm}
You are a fact-checking discriminator. Your task is to determine whether a candidate sentence explicitly contains or supports a given atomic claim.

Please analyze whether the candidate sentence explicitly contains or supports the atomic claim. The sentence should directly express or clearly support the information stated in the claim.

Respond with only "YES" if the sentence explicitly contains or supports the claim, or "NO" if it does not.

{\color{black}\bf Input:}\\
Atomic Claim: {\color{deepblue}\bf \{claim\}} \\
Candidate Sentence: {\color{deepblue}\bf \{sentence\}} \\
{\color{black}\bf Output:}\\

\end{AIbox}
\vspace{-1em}
\captionof{figure}{Prompt of Discriminator.}
\label{fig:discriminator}

\begin{AIbox}{Prompt of Matched-Budget LLM Routing Gate.}
{\color{black}\bf \large User Prompt:}
\tcblower
\vspace{1mm}
You are a routing gate for a fact-checking system. Given a single sentence, rate how factually COMPLEX it is. In other words, how much it would benefit from being decomposed into multiple interdependent atomic facts before verification.

Judge complexity using the following criteria.

HIGH complexity (score toward 1.0) if the sentence shows any of:
\begin{itemize}
    \item Multi-hop relations: verifying it requires chaining two or more facts.
    \item Several entities tied together: multiple named entities, dates, or quantities that are factually linked to each other, not merely listed.
    \item Compound or nested claims: coordinated clauses, subordinate clauses, or appositives that form several independent checkable points.
    \item High relation density: many explicit factual connections (predicative, attributive, temporal/spatial, or causal) among entity pairs, forming a connected, larger-diameter relation graph.
    \item Cross-clause coreference or context dependence: the checked target can only be resolved after coreference resolution.
\end{itemize}

LOW complexity (score toward 0.0) if the sentence is:
\begin{itemize}
    \item A single simple atomic statement (one subject-predicate-object fact).
    \item Built on few entities with no mutual binding (at most one core entity, or entities that need no joint verification).
    \item Free of nesting or coordination that would require splitting.
\end{itemize}

Return ONLY JSON:
\begin{lstlisting}[style=prompt]
{"complexity": <float between 0 and 1>}
\end{lstlisting}

Sentence: {\color{deepblue}\bf \{sentence\}}
\end{AIbox}
\vspace{-1em}
\captionof{figure}{Prompt used by the GPT-4o-mini and GPT-5 routing-gate baselines.}
\label{fig:routing-gate}

\section{Case Study}

\begin{AIbox}{Element Extraction and Verification Case.}
\small

\large \textbf{Question}
\begin{lstlisting}[style=prompt]
You should write a detailed response to the following question on food. [Question]: What will happen if I only eat vegetables for a week? [Requirements]: 1. The answer should be long enough to provide a comprehensive response.
\end{lstlisting}

\large \textbf{Answer}
\begin{lstlisting}[style=prompt]
Eating only vegetables for a week can lead to a mix of short-term benefits and potential challenges, depending on the variety, quantity, and preparation of the vegetables consumed. Below is a detailed analysis of the potential outcomes:

---

### **1. Nutrient Intake: Benefits and Gaps**
- **Positive Effects**: Vegetables are rich in vitamins (A, C, K, folate), minerals (potassium, magnesium), antioxidants, and fiber. A week of consuming a diverse range of vegetables (e.g., leafy greens, cruciferous veggies, root vegetables, and legumes) could boost your immune system, improve skin health, and support digestion due to the high fiber content. 
  - Example: Spinach and kale provide iron and calcium, while bell peppers and broccoli are high in vitamin C.
- **Potential Deficiencies**: 
  - **Protein**: Most vegetables (except legumes like lentils or chickpeas) are low in protein. A week without adequate protein may lead to muscle weakness, fatigue, or slower recovery from physical activity.
  - **Healthy Fats**: Vegetables lack essential fatty acids (e.g., omega-3s), which are critical for brain function and hormone production. Without fats, absorption of fat-soluble vitamins (A, D, E, K) may also be reduced.
  - **Vitamin B12 and D**: These are absent in plant-based foods (unless fortified), but a one-week deficiency is unlikely to cause harm. Longer-term exclusion would be problematic.

---

### **2. Digestive Changes**
- **Improved Gut Health**: The fiber in vegetables acts as a prebiotic, feeding beneficial gut bacteria. This might reduce bloating and improve bowel regularity over time.
- **Initial Discomfort**: A sudden increase in fiber (especially raw vegetables) can cause gas, bloating, or diarrhea if your gut isn't accustomed to it. Cooking vegetables or gradually increasing fiber intake can mitigate this.

---

### **3. Energy Levels and Metabolism**
- **Caloric Deficit**: Most vegetables are low in calories. If you don't eat enough volume to meet your energy needs, you may experience fatigue, headaches, or irritability. For example, 100g of spinach has only 23 calories, whereas 100g of potatoes provides 77 calories (and more starch for sustained energy).
- **Blood Sugar Stability**: Non-starchy vegetables (e.g., broccoli, zucchini) have minimal impact on blood sugar, which can help prevent energy crashes. However, relying solely on them without balancing with protein or fats might leave you feeling unsatisfied.

---

### **4. Weight Fluctuation**
- **Short-Term Weight Loss**: A vegetable-only diet often leads to water weight loss due to reduced sodium intake and lower carbohydrate stores (glycogen). However, this is temporary and not equivalent to fat loss.
- **Muscle Loss**: Insufficient protein and calories could cause the body to break down muscle for energy, especially if combined with physical activity.

---

### **5. Psychological and Social Impacts**
- **Mental Clarity**: Some people report improved focus and reduced brain fog due to the anti-inflammatory properties of vegetables.
- **Cravings and Mood Swings**: Restricting entire food groups (e.g., grains, fats, proteins) may trigger cravings for salty, sweet, or fatty foods. Social situations (e.g., dining out) could also become challenging.

---

### **6. Long-Term Considerations**
While a one-week vegetable-focused diet is unlikely to cause severe harm, prolonged restriction of other food groups could lead to nutrient deficiencies, hormonal imbalances (e.g., thyroid issues from excessive raw cruciferous vegetables), or disrupted menstrual cycles in women. 

---

### **Recommendations for a Balanced Approach**
If you plan to eat only vegetables for a week:
- **Prioritize Variety**: Include legumes (lentils, beans) for protein, starchy vegetables (sweet potatoes, squash) for energy, and avocado or olive oil for healthy fats.
- **Monitor Portions**: Ensure you're eating enough to meet daily caloric needs (~1,600-2,500 calories depending on activity level).
- **Cook Vegetables**: Steaming or roasting improves digestibility and nutrient absorption (e.g., lycopene in tomatoes is better absorbed when cooked).
- **Supplement Wisely**: Consider a B12 or vitamin D supplement if your diet lacks fortified foods.

---

### **Conclusion**
A week of eating only vegetables can be a refreshing "reset" that highlights the benefits of plant-based foods, such as increased fiber and micronutrient intake. However, it may also expose you to short-term energy dips, cravings, and digestive adjustments. For most people, this experiment is safe but should not be extended long-term without careful planning to avoid nutrient gaps. Always consult a healthcare professional before making significant dietary changes, especially if you have underlying health conditions.
\end{lstlisting}

\large \textbf{Verifiable Sentences }

 Sentence 1
\begin{lstlisting}[style=prompt]
{
  "Sentence": "Eating only vegetables for a week can lead to a mix of short-term benefits and potential challenges, depending on the variety, quantity, and preparation of the vegetables consumed.",
  "element": "pairs": [
        ["Eating only vegetables", "week"],
        ["Eating only vegetables", "mix"],
        ["mix", "short-term benefits"],
        ["mix", "potential challenges"],
        ["mix", "variety"],
        ["mix", "quantity"],
        ["mix", "preparation"],
        ["variety", "vegetables consumed"],
        ["quantity", "vegetables consumed"],
        ["preparation", "vegetables consumed"]
      ],
     "verification": {
        "support": [
        ["Eating only vegetables", "week"],
        ["Eating only vegetables", "mix"],
        ["mix", "short-term benefits"],
        ["mix", "potential challenges"],
        ["mix", "variety"],
        ["mix", "quantity"],
        ["mix", "preparation"],
        ["variety", "vegetables consumed"],
        ["quantity", "vegetables consumed"],
        ["preparation", "vegetables consumed"],
        "refute": [],
        "not_enough_evidence": [],
        "conflicting_evidence": []
      },
    "initial_verdict": "supported",
    "final_verdict": "supported",
}
\end{lstlisting}

\begin{center}
\textcolor{gray}{\Large $\cdots$}
\end{center}

 Sentence 3
\begin{lstlisting}[style=prompt]
{
  "sentence": "A week of consuming a diverse range of vegetables (e.g., leafy greens, cruciferous veggies, root vegetables, and legumes) could boost a person's immune system, improve skin health, and support digestion due to the high fiber content.",
  "pairs": [
    ["week", "diverse range of vegetables (e.g., leafy greens, cruciferous veggies, root vegetables, and legumes)"],
    ["week", "person's immune system"],
    ["week", "skin health"],
    ["week", "digestion"],
    ["digestion", "high fiber content"]
  ],
  "coref": [],
  "verification": {
    "support": [],
    "refute": [],
    "not_enough_evidence": [
      ["week", "diverse range of vegetables (e.g., leafy greens, cruciferous veggies, root vegetables, and legumes)"],
      ["week", "person's immune system"],
      ["week", "skin health"],
      ["week", "digestion"],
      ["digestion", "high fiber content"]
    ],
    "conflicting_evidence": []
  },
  "initial_verdict": "needs_reverify",
  "reverify_info": {
    "decomposed_claim": "A week of consuming a diverse range of vegetables (e.g., leafy greens, cruciferous veggies, root vegetables, and legumes) supports digestion due to the high fiber content.",
    "evidence": "Title: Eating 30 Plants per Week: How To Do It and Why - ZOE\nDate published: Nov 11, 2025\nSnippet:\nResearch suggests that eating 30 plant types every week could have a particularly positive effect on the diversity and health of your gut microbiome.\n\nTitle: Eating 30 Different Plants a Week: How to Do It - My Doctor Online\nDate published: Jun 10, 2024\nSnippet:\nA doctor explains the health benefits of getting 30 plant points a week and meal plan ideas.\n\nTitle: Cruciferous Vegetables: 5 Health Benefits\nDate published: No record\nSnippet:\nCruciferous vegetables include leafy greens, broccoli, Brussels sprouts and more. They have anti-cancer properties and are high in fiber and vitamins C, ...\n\nTitle: You Should Aim To Eat 30 Different Plants a Week - Food Network\nDate published: Apr 9, 2025\nSnippet:\nResearchers found that eating at least 30 different plants a week helped people diversify their microbiome.\n\nTitle: Why to eat cruciferous vegetables - Mayo Clinic Health System\nDate published: Oct 21, 2024\nSnippet:\nAdding cruciferous vegetables to your recipes is an easy way to increase fiber, support a healthy diet and reduce obesity.\n\nTitle: Diversity of Plant-Based Food Consumption: A Systematic Scoping ...\nDate published: Apr 29, 2025\nSnippet:\nHigher fruit and vegetable diversity was associated with higher vitamin C, carotenoids, flavonoids, and dietary fiber, compared with lower ...\n\nTitle: Health Benefits of Cruciferous Vegetables - WebMD\nDate published: Nov 14, 2024\nSnippet:\nLow in calories and high in fiber, cruciferous vegetables also help you feel fuller longer. That can help people eating for weight loss or heart ...\n\nTitle: The Power of Cruciferous Vegetables - Experience Life Magazine\nDate published: Jun 1, 2021\nSnippet:\nStudies have found that crucifers - especially one of their main compounds, sulforaphane - benefit metabolic and heart health ...\n\nTitle: 11 Reasons to Eat Vegetables Every Day | Orlando - UCF Health\nDate published: No record\nSnippet:\nVegetables",
    "reverify_label": "support"
  },
  "final_verdict": "supported"
}


\end{lstlisting}

\begin{center}
\textcolor{gray}{\Large $\cdots$}
\end{center}

 Sentence 21
\begin{lstlisting}[style=prompt]
{
  "sentence": "Insufficient protein and calories could cause the body to break down muscle for energy, especially if combined with physical activity.",
  "pairs": [
    ["Insufficient protein", "body"],
    ["calories", "body"],
    ["body", "muscle"],
    ["muscle", "energy"],
    ["Insufficient protein", "physical activity"],
    ["calories", "physical activity"]
  ],
  "coref": [],
  "verification": {
    "support": [
      ["Insufficient protein", "body"],
      ["calories", "body"],
      ["body", "muscle"],
      ["muscle", "energy"],
      ["calories", "physical activity"]
    ],
    "refute": [],
    "not_enough_evidence": [
      ["Insufficient protein", "physical activity"]
    ],
    "conflicting_evidence": []
  },
  "initial_verdict": "needs_reverify",
  "reverify_info": {
    "decomposed_claim": "Insufficient protein intake combined with physical activity can lead to muscle breakdown for energy.",
    "evidence": "Title: The effect of protein intake on athletic performance - PubMed Central\nDate published: Nov 6, 2024\nSnippet:\nProtein intake appears to provide modest benefits to athletes in improving their performance, particularly by enhancing endurance.\n\nTitle: Skeletal Muscle Responses to Negative Energy Balance: Effects of ...\nDate published: No record\nSnippet:\nDecreases in skeletal muscle mass in response to negative energy balance are due to imbalanced rates of muscle protein synthesis and degradation.\n\nTitle: Protein Intake and Exercise-Induced Skeletal Muscle Hypertrophy\nDate published: Jul 7, 2020\nSnippet:\nThe combination of protein intake and resistance exercise is the most efficient strategy to promote skeletal muscle hypertrophy and remodeling.\n\nTitle: Muscle preservation during hospitalization: energy balance, protein ...\nDate published: No record\nSnippet:\nPhysical inactivity, along with reduced energy and protein intake, are the primary drivers of muscle atrophy during hospitalization by suppressing muscle ...\n\nTitle: Low protein intake, muscle strength and physical performance in the ...\nDate published: No record\nSnippet:\nLow protein intake has been linked to reduced muscle strength and physical performance in older adults but little is known about how it may affect muscle ...\n\nTitle: Protein Intake & Muscle Mass in Elderly Females with Sarcopenia\nDate published: May 11, 2025\nSnippet:\nHowever, inadequate protein intake, along with hormonal imbalances, reduced physical activity, and chronic inflammation, accelerates muscle loss ...\n\nTitle: Protein & Energy Intake Not Linked to Muscle Mass in Older Adults\nDate published: No record\nSnippet:\nWe found that neither daily total protein intake nor distribution throughout the day were associated with muscle mass, strength, or physical function.\n\nTitle: Effects of protein supplementation on muscle mass, muscle strength ...\nDate published: Apr 8, 2025\nSnippet:\nA deficiency in protein intake could lead to an imbalance in pr",
    "reverify_label": "not enough evidence"
  },
  "final_verdict": "not enough evidence"
}

\end{lstlisting}
\begin{center}
\textcolor{gray}{\Large $\cdots$}
\end{center}

 Sentence 87
 \begin{lstlisting}[style=prompt]
{
  "sentence": "Vegetables do not contain complete proteins (complete proteins that include all nine essential amino acids), so a diet solely based on vegetables could lead to protein deficiency.",
  "pairs": [
    ["Vegetables", "complete proteins"],
    ["complete proteins", "nine essential amino acids"],
    ["diet", "vegetables"],
    ["diet", "protein deficiency"]
  ],
  "coref": [
    ["those", "complete proteins"]
  ],
  "verification": {
    "support": [
      ["complete proteins", "nine essential amino acids"]
    ],
    "refute": [
      ["diet", "protein deficiency"]
    ],
    "not_enough_evidence": [
      ["diet", "vegetables"]
    ],
    "conflicting_evidence": [
      ["Vegetables", "complete proteins"]
    ]
  },
  "initial_verdict": "refuted",
  "final_verdict": "refuted",
  "reverify_info": null
}


\end{lstlisting}

\end{AIbox}

\captionof{figure}{ElementCheck Case study.}
\label{fig:case-study}

\end{document}